\documentclass[10pt,twocolumn,letterpaper]{article}

\usepackage{lineno}
\usepackage[accsupp]{axessibility}

\usepackage{cvpr}              

\definecolor{cvprblue}{rgb}{0.21,0.49,0.74}
\usepackage[pagebackref,breaklinks,colorlinks,allcolors=cvprblue]{hyperref}
\usepackage{bm}
\usepackage[ruled,linesnumbered]{algorithm2e}
\usepackage{multirow}
\allowdisplaybreaks[4]

\def\paperID{12541} 
\def\confName{CVPR}
\def\confYear{2025}

\title{ReDiffDet: Rotation-equivariant Diffusion Model for Oriented Object Detection}

\author{
  Zeyu Ding$^{1,2}$ \ \  Yong Zhou  $^{1,2}$\thanks{\emph{Corresponding author}} \ \ Jiaqi Zhao$^{1,2}$ \ \ Hancheng Zhu$^{1,2}$ \ \ Wen-Liang Du$^{1,2}$ \ \ Rui Yao$^{1,2}$ \\
  $^1$  School of Computer Science and Technology, China University of Mining and Technology \\ 
  $^2$  Mine Digitization Engineering Research Center of the Ministry of Education \\
  \texttt{\normalsize \{dingzeyu,yzhou,jiaqizhao,zhuhancheng,wldu,ruiyao\}@cumt.edu.cn}  \\
}

\newcommand{\tablestyle}[2]{\setlength{\tabcolsep}{#1}\renewcommand{\arraystretch}{#2}\centering\footnotesize}

\usepackage{color, colortbl}

\newcommand\dunderline[2][.4pt]{%
  \raisebox{-#1}{\underline{\raisebox{#1}{\smash{\underline{#2}}}}}}
  
\newcommand{\notations}{
    \begin{table}[h]
    \centering
    \resizebox{0.9\linewidth}{!}
    {
        \begin{tabular}{l|l}
        \toprule
        \textbf{Notation} & \textbf{Description}    \\
        \midrule
        $\bm{z}_t$ & random variable                \\
        $t\in \{1,...,T\}$ & time steps             \\
        $q(\cdot |\cdot)$& diffusion process        \\
        $p_\theta(\cdot |\cdot)$& reverse process   \\
        $\boldsymbol{\mu}$ & mean                   \\
        $\boldsymbol{\Sigma}$ & variance            \\
        $\mathcal{N}(\boldsymbol{\mu},\space \boldsymbol{\Sigma})$ & Gaussian distribution    \\
        $\mathcal{N}(\mathbf{0}, \mathbf{I})$ & standard Gaussian distribution    \\
        $\mathbf{0}$ & zero matrix                  \\
        $\mathbf{I}$ & identity matrix              \\
        $(cx,cy)$ & center coordinate               \\
        $(w,h)$ & width and height                  \\
        $a$ & angle                                 \\
        $\mathbf{R}$ & rotation matrix              \\
        $\mathbf{\Lambda}$ & diagonal matrix        \\
        $SO(2)$       & 2D rotation group           \\
        $\mathbb{R}^2$ & 2D space                   \\
        $\mathrm{det}(\cdot)$    & determinant               \\
        $G$  & a group                              \\
        $\rho:X\to Y$           & a function        \\
        $S^{(\cdot)}_g$     & group action          \\
        $m$ & scaling factor                        \\
        $\exp(\cdot)$ & exponential function        \\
        $\mathrm{PDF}(\cdot,\cdot)$& probability density function \\
        $P$=$\{\mathbf{P_1},...,\mathbf{P}_k\}$ & samples \\
        $k$ & number of samples                     \\
        $(u_1,v_1)$ & a sample point                    \\
        $\lambda,\lambda_1,\lambda_2$ & eigenvalues \\
        $i$ & index                                 \\
        $\alpha$    & variance schedule             \\
        $\beta$     & variance schedule              \\
        $\sigma _t$ & variance schedule             \\
        $\epsilon_\theta(\bm{z}_{t},t)$ & model     \\
        $\theta$ & model parameter                  \\
        $\mathbb{E}(\cdot)$ & mean                  \\
        $\mathbb{V}(\cdot)$ & variance              \\
        \bottomrule
        \end{tabular}
    }
    \caption{The nomenclature with related notations.}
    \label{notations}
    \end{table}
}

\newcommand{\TableExpSetting}{
    \begin{table}[h]
    \centering
            \begin{tabular}{l|l}
            \toprule
             \textbf{Config}& \textbf{Value} \\ 
            \midrule
            optimizer      & AdamW       \\
            base learning rate  & 4e-5        \\
            weight decay        & 1e-4        \\
            optimizer momentum  & $\beta_{1},\beta_{2}$=0.9, 0.999 \\
            batch size          & 4 (2 images per GPU)   \\
            GPUs                & 2 NVIDA 2080ti     \\
            epochs              & 24    \\
            lr decay epochs     & (16, 22)     \\
            warmup iter         & 500         \\
            warmup factor       & 0.333       \\
            clip gradient type  & full model  \\
            clip gradient value & 1.0         \\
            clip gradiant norm  & 2.0         \\
            data augmentation   & only RandomFlip  \\ 
            seed                & Random Seed \\ 
            \hline
            \end{tabular}
        \caption{DIOR-R setting.}
            \label{expsetting}
    \end{table}
}

\newcommand{\DOTAOneResults}{
    \begin{table*}[t]   
    \centering
    \resizebox{1.0\textwidth}{!}
    {
        \begin{tabular}{lcccccccccccccccccc} 
        \toprule[1.2pt]
         &\textbf{Method} & \textbf{B.} & \textbf{PL} & \textbf{BD} & \textbf{BR} & \textbf{GTF}  & \textbf{SV} & \textbf{LV} & \textbf{SH} & \textbf{TC} & \textbf{BC} & \textbf{ST} & \textbf{SBF}   & \textbf{RA} & \textbf{HA} & \textbf{SP} & \textbf{HC} &\textbf{AP$_{50}$}\\ 
        \midrule[1pt]
          \multirow{8}{*}{\rotatebox{90}{\textbf{\textit{One-Stage}}}}&PSC~\cite{psc}      & R50   & 88.24 & 74.42 & 48.63 & 63.44 & 79.98 & 80.76 & 87.59 & 90.88 & 82.02 & 71.58 & 59.12 & 60.78 & 65.78 & 71.21 & 53.06 & 71.83 \\
          &R3Det~\cite{r3det}    & R101   & 88.76 & 83.09 & 50.91 & 67.27 & 76.23 & 80.39 & 86.72 & 90.78 & 84.68 & 83.24 & 61.98 & 61.35 & 66.91 & 70.63 & 53.94 & 73.79 \\
          &S2A-Net~\cite{s2anet}      & R50  & 89.11 & 82.84 & 48.37 & 71.11 & 78.11 & 78.39 & 87.25 & 90.83 & 84.90 & 85.64 & 60.36 & 62.60 & 65.26 & 69.13 & 57.94 & 74.12 \\
          &H2RBox~\cite{h2rbox} & R50   &88.93 & 78.89 & 46.27 & 68.79 & \dunderline{81.12} & 75.45 & 86.68 & 90.89 & 86.71 & 87.33 & 64.15 & 68.83 & 62.81 & 69.39 & 59.79 & 74.40 \\
          &CFA~\cite{cfa}      & R50   & 88.34 & 83.09 & 51.92 & 72.23 & 79.95 & 78.68 & 87.25 & 90.90 & 85.38 & 85.71 & 59.63 & 63.05 & 73.33 & 70.36 & 47.86 & 74.51 \\
          &DHRec~\cite{dhrec}   & R50  & 88.58 & 77.90 & 53.84 & 72.93 & 78.45 & 78.84 & 87.64 & 90.88 & \textbf{88.78} & 85.46 & 56.11 & 66.74 & 67.58 & 70.25 & 57.53 & 74.57 \\
          &SASM~\cite{sasm}     & R50   & 86.42 & 78.97 & 52.47 & 69.84 & 77.30 & 75.99 & 86.72 & 90.89 & 82.63 & 85.66 & 60.13 & 68.25 & 73.98 & 72.22 & 62.37 & 74.92 \\
          &FRED~\cite{fred} & Re50 & 89.37 & 82.12 & 50.84 & 73.89 & 77.58 & 77.38 & 87.51 & 90.82 & 86.30 & 84.25 & 62.54 & 65.10 & 72.65 & 69.55 & 63.41 & 75.56 \\
        \midrule[1pt]
          \multirow{6}{*}{\rotatebox{90}{\textbf{\textit{Two-Stage}}}}&RoI Transformer~\cite{roi_transformer}      & R50   & 88.65 & 82.60 & 52.53 & 70.87 & 77.93 & 76.67 & 86.87 & 90.71 & 83.83 & 52.81 & 53.95 & 67.61 & 74.67 & 68.75 & 61.03 & 74.61 \\
          &Gliding Vertex~\cite{gliding_vertex} & R101   & \dunderline{89.64} & \dunderline{85.00} & 52.26 & \dunderline{77.34} & 73.01 & 73.14 & 86.82 & 90.74 & 79.02 & 86.81 & 59.55 & \textbf{70.91} & 72.94 & 70.86 & 57.32 & 75.02 \\
          &AOPG~\cite{dior} & R50 & 89.27 & 83.49 & 52.50 & 69.97 & 73.51 & 82.31 & 87.95 & 90.89 & 87.64 & 84.71 & 60.01 & 66.12 & 74.19 & 68.30 & 57.80 & 75.24 \\
          &DODet~\cite{dodet}  & R50 & 89.34 & 84.31 & 51.39 & 71.04 & 79.04 & 82.86 & 88.15 & 90.90 & 86.88 & 84.91 & 62.69 & 67.63 & 75.47 & 72.22 & 45.54 & 75.49 \\
          &Oriented R-CNN~\cite{orientedrcnn} & R50   & 89.46 & 82.12 & 54.78 & 70.86 & 78.93 & 83.00 & 88.20 & 90.90 & 87.50 & 84.68 & 63.97 & 67.69 & 74.94 & 68.84 & 52.28 & 75.87 \\
          &SCRDet++ *~\cite{SCRDetplusplus}     & R101 & \textbf{89.77} & 83.90 & \dunderline{56.30} & 73.98 & 72.60 & 75.63 & 82.82 & 90.76 & 87.89 & 86.14 & \dunderline{65.24} & 63.17 & 76.05 & 68.06 & \dunderline{70.24} & 76.20 \\
        \midrule[1pt]
          \multirow{4}{*}{\rotatebox{90}{\textbf{\textit{Query}}}}&D. DETR-O~\cite{deformabledetr}     & R50   & 84.89 & 70.71 & 46.04 & 61.92 & 73.99 & 78.83 & 87.71 & 90.07 & 77.97 & 78.41 & 47.07 & 54.48 & 66.87 & 67.66 & 55.62 & 69.48 \\
          &EMO2-DETR~\cite{emo2detr}&R50 & 88.08 & 77.91 & 43.17 & 62.91 & 74.01 & 75.09 & 79.21 & 90.88 & 81.50 & 54.04 & 51.92 & 59.44 & 64.74 & 71.81 & 58.96 & 70.91  \\
          &ARS-DETR~\cite{arsdetr} & R50   & 86.97 & 75.56 & 48.32 & 69.20 & 77.92 & 77.94 & 87.69 & 90.50 & 77.31 & 82.86 & 60.28 & 64.58 & 74.88 & 71.76 & 66.62 & 74.16 \\
          &OrientedFormer~\cite{orientedformer}& R50  & 88.14 & 79.13 & 51.96 & 67.34 & 81.02 & 83.26 & 88.29 & 90.90 & 85.57 & 86.25 & 60.84 & 66.36 & 73.81 & 71.23 & 56.49 & 75.37 \\
         \midrule[1pt]
         \multirow{5}{*}{\rotatebox{90}{\textbf{\textit{Diffusion}}}}&DiffusionDet-O~\cite{diffusiondet} & R50  & 78.29 & 68.37 & 38.08 & 54.63 & 60.42 & 54.88 & 64.83 & 89.99 & 76.40 & 75.44 & 39.19 & 58.15 & 42.46 & 56.98 & 49.39 & 60.50\\
         &\textbf{\textit{ReDiffDet (ours)}} & ReR50 & 86.32 & 78.72 & 51.56 & 71.89 & 80.05 & 84.73 & 88.75 & 90.84 & 85.92 & 86.69 & 59.48 & 57.74 & 75.19 & 68.43 & 65.37 & 75.45\\
         &\textbf{\textit{ReDiffDet (ours)}} & R50 & 85.45 & 77.84 & 50.39 & 71.63 & 80.90 & 84.85 & 88.77 & 90.88 & 87.00 & 86.08 & 63.30 & 63.43 & 76.01 & \dunderline{76.03} & 60.16 & 76.18\\
         &\textbf{\textit{ReDiffDet (ours)}} & Swin-T  & 82.89 & 82.90 & 52.47 & 70.80 & 80.48 & \textbf{85.41} & \textbf{88.97} & \textbf{90.90} & 86.97 & \dunderline{87.35} & 60.24 & 65.01 & \dunderline{76.61} & 74.28 & 64.07 & \dunderline{76.62}\\
         &\textbf{\textit{  ReDiffDet (ours)*}} & ReR50  & 87.93 & \textbf{86.04} & \textbf{57.39} & \textbf{82.01} & \textbf{81.36} & \dunderline{84.92} & \dunderline{88.81} & \dunderline{90.89} & \dunderline{88.57} & \textbf{87.55} & \textbf{70.01} & \dunderline{69.99} & \textbf{79.80} & \textbf{80.93} & \textbf{74.26} & \textbf{80.70}\\
         \bottomrule[1.2pt]
        \end{tabular}
    }
    \vspace{-0.5em}
    \caption{Comparison with state-of-the-art methods on \textbf{DOTA-v1.0}~\cite{dotav1.0} dataset. * denotes multi-scale traning and testing. B. denotes the backbone. The \textbf{best} and \dunderline{second-best} results are highlighted in \textbf{bold} and \dunderline{underlined}.}
    \vspace{-1em}
    \label{tab:dota1.0}
    \end{table*}
}

\newcommand{\TableDOTAVonefiveandtwoResult}{
    \begin{table}[h]
    \tablestyle{3pt}{0.99}
    \centering
    \resizebox{\linewidth}{!}
    {
    \begin{tabular}{ccccc}
    \toprule[1pt]
    \textbf{Method} & R. Net-O~\cite{focal_loss} & M. R-CNN~\cite{mask_rcnn} & HTC~\cite{htc} & RoI Trans.~\cite{roi_transformer} \\
    \textbf{backbone}   &R50 &R50 &R50 &R50 \\
    \textbf{DOTA-v1.5}    & 59.16  & 62.67 & 63.40 & 65.50      \\ 
    \textbf{DOTA-v2.0}    & 46.68  & 49.47 & 50.34 & 52.81      \\ 
    \midrule[1pt]
    \textbf{Method} & ReDet~\cite{redet} & O. Former~\cite{orientedformer} & DCFL~\cite{dcfl} & \textbf{\textit{ReDiffDet (ours)}}  \\
    \textbf{backbone}   &ReR50 &R50 &R50 &ReR50 \\
    \textbf{DOTA-v1.5}    & 66.86 & 67.06 & 67.37 & \textbf{68.00}      \\ 
    \textbf{DOTA-v2.0}    & - & 54.27 & 51.57 & \textbf{55.76}      \\ 
    \bottomrule[1pt]
    \end{tabular}
    }
    \vspace{-0.5em}
    \caption{Main results with AP$_{50}$ on \textbf{DOTA-v1.5} and \textbf{DOTA-v2.0}.}
    \vspace{-0.5em}
    \label{tab:dota1.5and2.0}
    \end{table}
}

\newcommand{\DIORRResults}{
    \begin{table}[h]
    \centering
    \tablestyle{3.5pt}{1.1}
    \resizebox{\linewidth}{!}
    {
    \begin{tabular}{ccccccc}
    \toprule[1pt]
    \textbf{Method} & {\small \textbf{backbone}} &\textbf{APL} & \textbf{SH} & \textbf{BR} & \textbf{VE} & \textbf{AP$_{50}$}\\ 
    \midrule[0.8pt]
    RetinaNet-O~\cite{focal_loss} & R50  & 61.49 & 77.71 & 23.98 & 38.01 &   57.55\\
    Oriented Rep\cite{oriented_reppoints} & R50 &  70.03 & 87.82 & 39.14 & 51.57 &  66.71\\
    DCFL~\cite{dcfl}  & R50 & 68.60 & 85.20  & 42.10 & 50.90 &   66.80\\
    Gliding Vertex~\cite{gliding_vertex} & R50 & 65.35 & 80.22 & 33.88 & 47.71 &   60.06\\
    RoI Transformer~\cite{roi_transformer}  & R50 & 63.34 & 81.21 & 40.68 & 43.27 &  63.87\\
    AOPG~\cite{dior}  & R50 & 62.39 & 81.17 & 40.90 & 52.38 &  64.41\\
    ARS-DETR~\cite{arsdetr}  & R50 & 68.00 & 84.55 & 41.13 & 50.57 &  66.12\\
    OrientedFormer~\cite{orientedformer} & R50 & 65.65 & 88.66 & 41.90 & 56.54 &  \dunderline{67.28}\\ 
    DiffusionDet-O~\cite{diffusiondet}  & R50 & 58.84 & 53.01 & 21.20 & 37.44 &  49.91\\
    \textbf{\textit{ReDiffDet (ours)}}  & PKINet-T & 66.69 & 88.78 & 42.33 & 56.39 &  65.47\\
    \textbf{\textit{ReDiffDet (ours)}}  & LSK-T & \dunderline{70.81} & \dunderline{89.27} & \dunderline{43.66} & \textbf{58.22} &  66.62\\
    \textbf{\textit{ReDiffDet (ours)}}  & ReR50 & \textbf{71.36} & \textbf{89.52} & \textbf{47.12} & \dunderline{57.32} &  \textbf{68.05}\\
    \bottomrule[1pt]
    \end{tabular}
    }
    \vspace{-0.5em}
    \caption{Main results of small size objects with AP$_{50}$ on \textbf{DIOR-R}.}
    \label{tab:diorr}
    \end{table}
}

\newcommand{\COCOResults}{
    \begin{table}[h]
    \centering
    \resizebox{\linewidth}{!}{
        \begin{tabular}{cccccccc}
        \toprule[1pt]
        \textbf{Method} & \textbf{mAP} & \textbf{AP$_{50}$} & \textbf{AP$_{75}$} & \textbf{AP$_{s}$} & \textbf{AP$_{m}$} & \textbf{AP$_{l}$}  \\ 
        \midrule[0.8pt]
        DiffusionDet~\cite{diffusiondet} & 43.2 & 62.6 & 46.1 & 25.4 & 46.0 & 59.2 \\
        \rowcolor{gray!15} & 45.0 & 64.6 & 48.4 & 27.0 & 47.4 & 59.9 \\ 
        \rowcolor{gray!15} \multirow{-2}{*}{+ Condition} & \textbf{\textcolor[rgb]{0,0.6,0}{\small +1.8}} & \textbf{\textcolor[rgb]{0,0.6,0}{\small +2.0}} & \textbf{\textcolor[rgb]{0,0.6,0}{\small +2.3}} & \textbf{\textcolor[rgb]{0,0.6,0}{\small +1.6}} & \textbf{\textcolor[rgb]{0,0.6,0}{\small +1.4}} & \textbf{\textcolor[rgb]{0,0.6,0}{\small +0.7}} \\
        \bottomrule[1pt]
        \end{tabular}
    }
    \vspace{-0.5em}
        \caption{Main results across various AP metrics on \textbf{COCO}~\cite{coco}.}
    \vspace{-1em}
    \label{tab:coco}
    \end{table}
}

\newcommand{\TableAblationStudiesIntegrated}{
\begin{table*}[!t]
\centering
        \subfloat[
        \textbf{Decoder layer}. 4 layers work best.
        \label{tab:decoder_layer}
        ]{
        \centering
        \begin{minipage}{0.16\linewidth}{\begin{center}
        \tablestyle{2pt}{1.26}
        \resizebox{1.15\textwidth}{!}{
            \begin{tabular}{cccc}
            \hline
             \addlinespace[1pt]
            \multicolumn{1}{c}{\textbf{Layer}}  & \textbf{Params} & \textbf{AP$_{75}$} & \textbf{AP$_{50}$} \\
            \addlinespace[1pt]
            \hline
            \multicolumn{1}{c}{3} & 51M & 49.14 & 67.25 \\
            \rowcolor{gray!15}\multicolumn{1}{c}{4} & 64M & 49.96 & 68.05 \\
            \multicolumn{1}{c}{5} & 78M & 49.65 & 67.74 \\
            \multicolumn{1}{c}{6} & 92M & 48.72 & 66.83 \\
            \multicolumn{1}{c}{7} & 106M & 48.16 & 66.25 \\
            \hline
            &&&\\
            \end{tabular}
        }
        \end{center}}
        \end{minipage}
        }
\hspace{1.1em}
        \subfloat[
        \textbf{Scaling factor $m$}. $m = 6$ works best.
        \label{tab:scaling_factor}
        ]{
        \centering
            \begin{minipage}{0.16\linewidth}
                {\begin{center}
                \tablestyle{2.5pt}{1.19}
                \resizebox{1.15\textwidth}{!}{
                    \begin{tabular}{cccc}
                    \hline
                    \addlinespace[1pt]
                    \multicolumn{1}{c}{$m$} & \textbf{AP{\tiny$_{50:95}$}} & \textbf{AP$_{75}$} & \textbf{AP$_{50}$} \\
                    \addlinespace[1pt]
                    \hline
                        \multicolumn{1}{c}{1.0} & 45.59 & 48.54 & 66.72  \\
                        \multicolumn{1}{c}{2.0} & 45.79 & 48.67 & 66.81 \\
                        \multicolumn{1}{c}{4.0} & 46.16 & 49.16 & 67.26 \\
                        \rowcolor{gray!15}\multicolumn{1}{c}{6.0} & 46.59 & 49.96 & 68.05 \\
                        \multicolumn{1}{c}{8.0} & 45.91 & 48.95 & 67.08 \\
                    \hline
                    &&&\\
                    \end{tabular}
                }
                \end{center}}
            \end{minipage}
        }
\hspace{1.1em}
        \subfloat[
        \textbf{Sample} {\scriptsize $\{\mathbf{P}_1,...,\mathbf{P}_k\}$}. $k = 8$ works best.
        \label{tab:samples}
        ]{
        \centering
            \begin{minipage}{0.16\linewidth}
                {\begin{center}
                \tablestyle{2pt}{1.16}
                \resizebox{1.15\textwidth}{!}{
                    \begin{tabular}{cccc}
                    \hline
                    \addlinespace[1pt]
                    \multicolumn{1}{c}{$k$} & \textbf{AP{\tiny$_{50:95}$}} & \textbf{AP$_{75}$} & \textbf{AP$_{50}$} \\
                    \addlinespace[1pt]
                    \hline
                    \multicolumn{1}{c}{4} & 46.44& 49.72 & 67.80  \\
                    \rowcolor{gray!15}\multicolumn{1}{c}{8} & 46.59 & 49.96 & 68.05 \\
                    \multicolumn{1}{c}{32} & 46.31& 49.51 & 67.61 \\
                    \multicolumn{1}{c}{64} & 46.27& 49.45 & 67.50 \\
                    \multicolumn{1}{c}{128} & 46.24& 49.29 & 67.37 \\
                    \hline
                     &&&\\
                    \end{tabular}
                }
                \end{center}}
            \end{minipage}
        }
\hspace{1.1em}
        \subfloat[
        \textbf{Signal scale}. The best signal scale is 2.0.
        \label{tab:signal_scale}
        ]{
        \centering
            \begin{minipage}{0.16\linewidth}
                {\begin{center}
                \tablestyle{2pt}{1.19}
                \resizebox{1.17\textwidth}{!}{
                    \begin{tabular}{cccc}
                    \hline
                    \addlinespace[1pt]
                    \multicolumn{1}{c}{\textbf{SNR}} & \textbf{AP{\tiny$_{50:95}$}} & \textbf{AP$_{75}$} & \textbf{AP$_{50}$} \\
                    \addlinespace[1pt]
                    \hline
                    \multicolumn{1}{c}{0.1} & 45.83& 48.83 & 66.87  \\
                    \multicolumn{1}{c}{1.0} & 45.97& 49.06 & 67.17  \\
                    \rowcolor{gray!15}\multicolumn{1}{c}{2.0} & 46.59  & 49.96 & 68.05 \\
                    \multicolumn{1}{c}{3.0} & 46.40& 49.67 & 67.77  \\
                    \multicolumn{1}{c}{4.0} & 45.65& 48.63 & 66.75 \\
                    \hline
                    &&&\\
                    \end{tabular}
                }
                \end{center}}
            \end{minipage}
        }
\hspace{1.1em}
        \subfloat[
        \textbf{Individual modules}.
        \label{tab:individual}
        ]{
        \centering
            \begin{minipage}{0.16\linewidth}
                {\begin{center}
                \tablestyle{2pt}{1.06}
                \resizebox{1.1\textwidth}{!}{
                    \begin{tabular}{cccc}
                    \hline
                    \addlinespace[1pt]
                    \textbf{Diff.} & \textbf{Enc.} & \textbf{Dec.} & \textbf{AP$_{50}$} \\
                    \addlinespace[1pt]
                    \hline
                                    &            &            & 49.91 \\
                                    &            & \checkmark & 52.15 \\
                                    & \checkmark &            & 60.02 \\
                                    & \checkmark & \checkmark & 62.17 \\
                         \checkmark & \checkmark &            & 65.59 \\
                    \rowcolor{gray!15} \checkmark & \checkmark & \checkmark & 68.05 \\
                    \hline
                    \end{tabular}
                }
                \end{center}}
            \end{minipage}
        }
        \vspace{-0.5em}
        \caption{\textbf{ReDiffDet ablation experiments} with ReResNet-50 on DIOR-R. The iteration step is 1 and number of boxes is 300 by default in this ablation. We report AP$_{50}$, AP$_{75}$, AP$_{50:95}$, and parameters.}
        \vspace{-1.5em}
\end{table*}
}

\newcommand{\TableParamsandFPS}{
    \begin{table}[!t]
    \centering
    \tablestyle{2pt}{1}
    \resizebox{\linewidth}{!}
    {
    \begin{tabular}{ccccc}
    \toprule[0.99pt]
    \textbf{Method} & O. R-CNN~\cite{orientedrcnn} & D. DETR-O~\cite{deformabledetr} & ARS-DETR~\cite{arsdetr} & O. Former~\cite{orientedformer} \\
    \textbf{backbone}   &R50 &R50 &R50 &R50 \\
    \textbf{Params, FPS}    & 41M, 8  & 41M, 11 & 42M, 10 & 44M, 11      \\ 
    \midrule[0.99pt]
    \textbf{Method} & DiffusionDet-O~\cite{diffusiondet} & \textbf{\textit{ReDiffDet}} & \textbf{\textit{ReDiffDet}} & \textbf{\textit{ReDiffDet}}  \\
    \textbf{backbone}   &R50 &Swin-T & R50 & ReR50 \\
    \textbf{Params, FPS}    & 84M, 15  & 91M, 14 & 88M, 14 & 64M, 14      \\ 
    \bottomrule[0.99pt]
    \end{tabular}
    }
    \caption{Comparison of \textbf{Params} and \textbf{FPS} with 1024$\times$1024 inputs.}
    \label{tab:paramsandfps}
    \vspace{-0.5em}
    \end{table}
}

\newcommand{\Algorithmtrain}{ 
    \begin{algorithm}[t]
        \small
        \SetAlgoLined
        \DontPrintSemicolon
        \caption{Training}
        \label{algorithmtrain}
        \KwIn{image, gt$\_$box, gt$\_$label, $T$}
        \KwOut{loss}       
        feat = backbone$\_$with$\_$FPN(image)
        
        con$\_$box = conditional$\_$encoder(feat)
        
        gt$\_$Gau = 2D$\_$Gaussian(gt$\_$box)
        
        pad$\_$Gau = pad(gt$\_$Gau, 2D$\_$Gaussian(con$\_$box))
        
        $P$ = sample(pad$\_$Gau)
        
        $\bm z_0$ = transform($P$) * scale
        
        $t$ = randint(0, $T$); $\epsilon \in \mathcal{N}(\mathbf{0}, \mathbf{I})$
        
        crpt$\_$pts = $\sqrt{\bar{\alpha}_t}\bm z_0+\sqrt{1-\bar{\alpha}_t}\epsilon$
        
        crpt$\_$Gau = MLE(crpt$\_$pts)
                
        pred$\_$box, pred$\_$label = aligned$\_$decoder(crpt$\_$Gau, feat, $t$)
        
        loss = set$\_$loss((pred$\_$box, gt$\_$box), (pred$\_$label, gt$\_$label))
        
        \Return loss    
    \end{algorithm}
}

\newcommand{\Algorithminference}{ 
    \begin{algorithm}[t]
        \small
        \SetAlgoLined
        \DontPrintSemicolon
        \caption{Inference}
        \label{algorithminference}
        \KwIn{image, step, $T$}
        \KwOut{box, label}
        \SetKw{KwIn}{in}
        feat = backbone$\_$with$\_$FPN(image)
        
        con$\_$box = conditional$\_$encoder(feat)
        
        Gau$\_$\textit{t} = transform(normal(mean=$\mathbf{0}$, std=$\mathbf{I}$), con$\_$box)
        
        time = reversed(linesapce(-1, $T$, step))
        
        time$\_$pair = list(zip(time[:-1], time[1:]))
        
        \For{t, t$\_$next \KwIn \textup{zip(time$\_$pair)}}{
        
            box, label = aligned$\_$decoder(Gau$\_$\textit{t}, feat, \textit{t})
            
            Gau$\_$\textit{pred} = 2D$\_$Gaussian(box)
            
            Gau$\_$\textit{t} = ddim$\_$step(Gau$\_$\textit{t}, Gau$\_$\textit{pred}, \textit{t}, \textit{t$\_$next})
            
            Gau$\_$\textit{t} = Gau$\_$renewal(Gau$\_$\textit{t})
        }
         
        \Return box, label
    \end{algorithm}
}

\newcommand{\DIORRResultsappendix}{
    \begin{table*}[h]
    \setlength{\tabcolsep}{2pt}
    \renewcommand\arraystretch{1.15}
    \scriptsize      
    \centering
    \resizebox{\textwidth}{!}
    {
        \hspace{-1.0em}
        \begin{tabular}{r|c|cccccccccccccccccccc|l}
        \toprule
         \textbf{Method} & \textbf{backbone}  & \textbf{APL}  & \textbf{APO}  & \textbf{BF}   & \textbf{BC}   & \textbf{BR}   & \textbf{CH}   & \textbf{DAM}  & \textbf{ETS}  & \textbf{ESA}  & \textbf{GF}   & \textbf{GTF}  & \textbf{HA}   & \textbf{OP}   & \textbf{SH}   & \textbf{STA}  & \textbf{STO}  & \textbf{TC}   & \textbf{TS}   & \textbf{VE}   & \textbf{WM}   & \textbf{AP$_{50}$}  \\ 
        \midrule
        \multicolumn{2}{l}{\textbf{\textit{One-stage}}}\\
        \midrule
        RetinaNet-O~\cite{focal_loss} & ResNet50   & 61.49    & 28.52    & 73.57    & 81.17    & 23.98    & 72.54    & 19.94    & 72.39    & 58.20    & 69.25    & 79.54    & 32.14    & 44.87    & 77.71    & 67.57    & 61.09    & 81.46    & 47.33    & 38.01    & 60.24    & 57.55    \\
        Oriented Rep~\cite{oriented_reppoints} & ResNet50  & 70.03 & 46.11    & 76.12    & 87.19    & 39.14    & 78.76    & 34.57    & 71.80    & 80.42    & 76.16    & 79.41    & 45.48    & 54.90    & 87.82    & 77.03    & 68.07    & 81.60    & 56.83    & 51.57    & \textbf{71.25} & 66.71    \\
        DCFL~\cite{dcfl} & ResNet50  & 68.60    & 53.10    & 76.70 & 87.10    & 42.10    & 78.60 & 34.50    & 71.50    & 80.80    & \textbf{79.70} & 79.50    & 47.30    & \textbf{57.40}    & 85.20    & 64.60    & 66.40    & 81.50    & 58.90    & 50.90    & 70.90    & 66.80    \\
        \midrule
        \multicolumn{2}{l}{\textbf{\textit{Two-stage}}}\\
        \midrule
        Gliding Vertex~\cite{gliding_vertex} & ResNet50   & 65.35    & 28.87    & 74.96    & 81.33    & 33.88    & 74.31    & 19.58    & 70.72    & 64.70    & 72.30    & 78.68    & 37.22    & 49.64    & 80.22    & 69.26    & 61.13    & 81.49    & 44.76    & 47.71    & 65.04    & 60.06    \\
        RoI Transformer~\cite{roi_transformer} & ResNet50 & 63.34    & 37.88    & 71.78    & 87.53    & 40.68    & 72.60    & 26.86    & \textbf{78.71}    & 68.09    & 68.96    & \textbf{82.74}    & \textbf{47.71}    & 55.61    & 81.21    & \textbf{78.23} & 70.26    & 81.61    & 54.86    & 43.27    & 65.52    & 63.87    \\
        AOPG~\cite{dior} & ResNet50  & 62.39    & 37.79    & 71.62    & 87.63    & 40.90    & 72.47    & 31.08    & 65.42    & 77.99    & 73.20    & 81.94    & 42.32    & 54.45    & 81.17    & 72.69    & 71.31    & 81.49    & 60.04 & 52.38    & 69.99    & 64.41    \\
        \midrule
        \multicolumn{2}{l}{\textbf{\textit{End-to-End}}}\\
        \midrule
        ARS-DETR~\cite{arsdetr} & ResNet50  & 68.00    & \textbf{54.17}    & 74.43    & 81.65    & 41.13    & 75.66    & \textbf{34.89}    & 73.07    & 81.92 & 76.10 & 78.62  & 36.33    & 55.41    & 84.55    & 70.09    & 72.23    & 81.14    & \textbf{61.52}    & 50.57    & 70.28    & 66.12    \\
        OrientedFormer~\cite{orientedformer} & ResNet50 & 65.65    & 48.69 & \textbf{78.79}    & 87.17    & 41.90 & 76.34    & 34.37 & 72.14    & 81.40    & 75.34    & 79.83    & 45.15    & 56.12 & 88.66 & 67.59    & 72.68 & 87.32    & 60.31    & 56.54 & 69.56    &67.28 \\
        \midrule
        \multicolumn{2}{l}{\textbf{\textit{Diffusion Model}}}\\
        \midrule
        DiffusionDet-O~\cite{diffusiondet} & ResNet50 & 58.84 & 24.25 & 70.10 & 78.93 & 21.20 & 72.25 & 21.93 & 53.19 & 53.82 & 56.26 & 74.26 & 1.71 & 37.82 & 53.01 & 62.66 & 50.48 & 80.36 & 27.92 & 37.44 & 61.80 & 49.91 \\
        \rowcolor{gray!15}ReDiffDet (ours) & PKINet-T & 66.69 & 42.46 & 77.02 & 84.74 & 42.33 & 74.14 & 30.06 & 69.15 & 79.12 & 71.03  & 79.76 & 39.57 & 56.44 & 88.78 & 73.37 & 75.72 & 85.26 & 53.47 & 56.39 & 63.92 & 65.47\\
        \rowcolor{gray!15}ReDiffDet (ours) & LSK-T & 70.81 & 44.08 & 76.03 & 85.29 & 43.66 & 75.90 & 31.81 & 71.83 & 81.75 & 71.52 & 81.42 & 41.82 & 56.91 & 89.27 & 75.25 & 75.32 & 87.42 & 54.43 & \textbf{58.22} & 59.69 & 66.62 \\
        \rowcolor{gray!15}ReDiffDet (ours) & ReR50 & \textbf{71.36} & 49.22 & 71.65 & \textbf{87.88} & \textbf{47.12} & \textbf{79.28} & 33.35 & 73.37 & \textbf{83.74} & 70.29 & 80.38 & 43.63 & 57.17 & \textbf{89.52} & 72.39 & \textbf{79.81} & \textbf{89.03} & 57.34 & 57.32 & 67.23 & \textbf{68.05} \\
        \bottomrule
        \end{tabular}
    }
    \caption{Experimental results on \textbf{DIOR-R} dataset.}
    \label{tab:diorrappendix}
    \end{table*}
}

\usepackage{graphicx}

\newcommand{\teaser}{
    \begin{figure}[t]
      \centering
       \includegraphics[width=\linewidth]{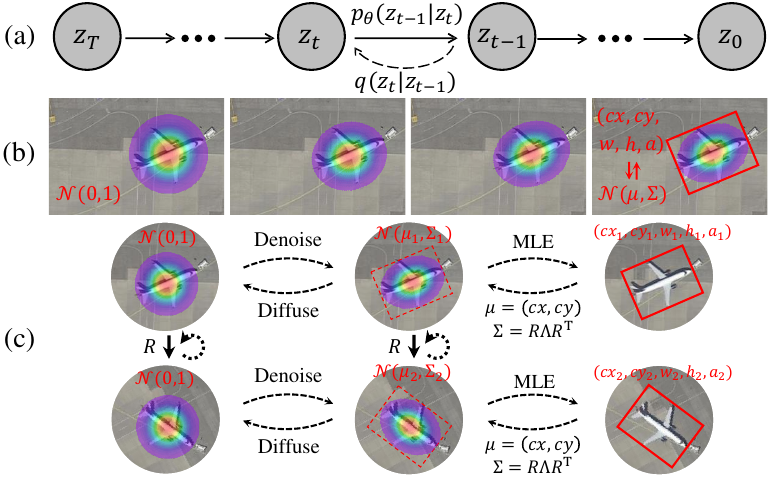}
       \caption{(a) $q(\cdot)$ is the diffusion process and $p_{\theta}(\cdot)$ is the reverse process. (b) ReDiffDet formulate oriented object detection as a denoising diffusion process from standard normal distribution to gaussian distribution of oriented boxes. (c) The rotation-equivariant diffusion model in ReDiffDet.}
       \label{teaser}
    \end{figure}
}

\newcommand{\analysisinintro}{
    \begin{figure*}[t]
      \centering
       \includegraphics[width=0.99\textwidth]{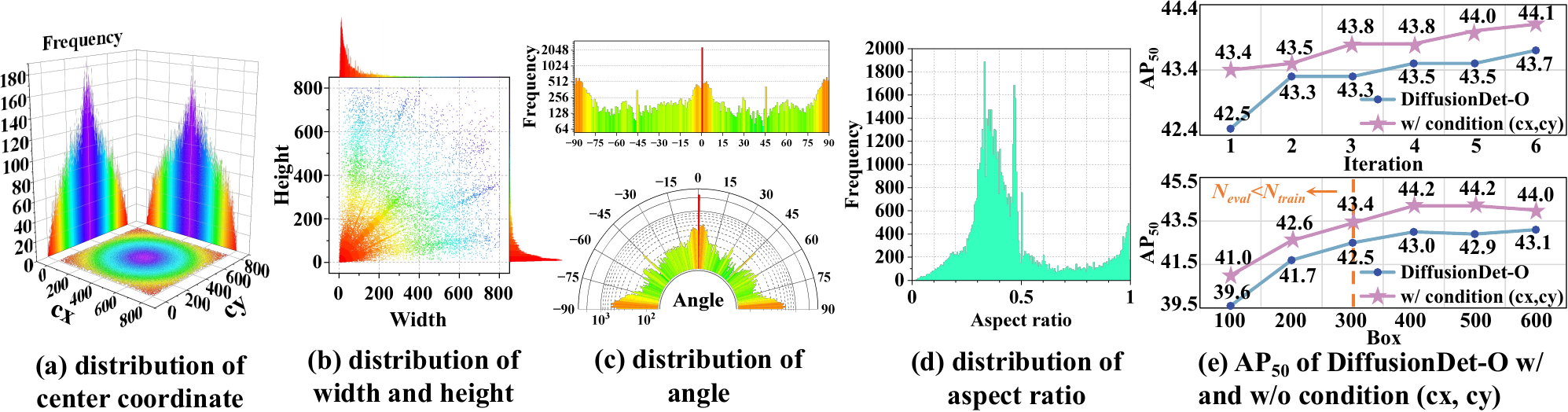}
       \caption{Distribution of oriented box characteristics on DIOR-R, including (a) center coordinates, (b) width and height, (c) angle, and (d) aspect ratio. (e) AP$_{50}$ comparison of DiffusionDet-O with and without conditional center coordinates on DIOR-R over 12 epochs.}
       \label{analysisinintro}
       \vspace{-0.7em}
    \end{figure*}
}

\newcommand{\overview}{
    \begin{figure*}[t]
      \centering
       \includegraphics[width=0.99\textwidth]{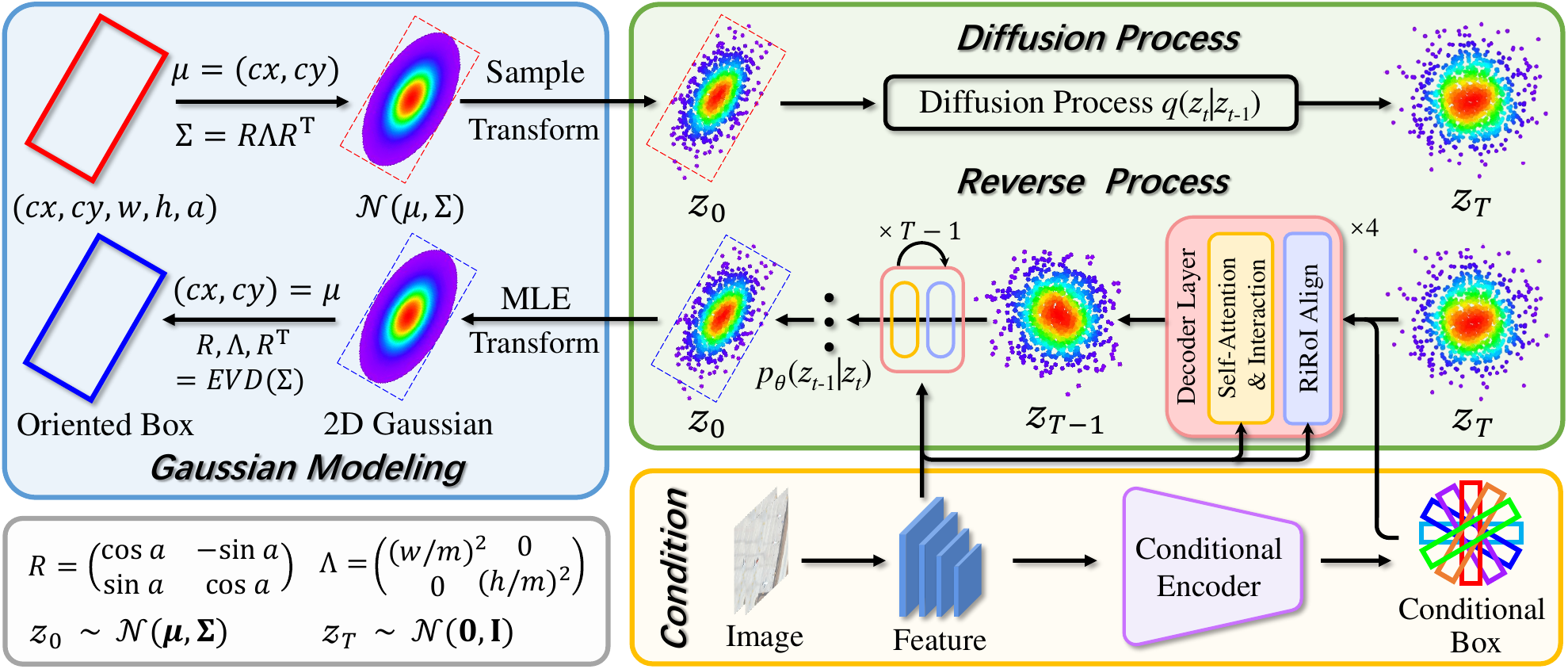}
       \caption{\textbf{Overview of ReDiffDet.} We represent an oriented box as a 2D Gaussian distribution. Gaussian distributions of boxes diffuse into standard normal distribution in the diffusion process. ReDiffDet generates boxes using the reverse process in inference. The conditional encoder provides conditional boxes. The aligned decoder layer consists of RiRoI Align, self-attention, and instance interaction.}
       \label{overview}
       \vspace{-0.5em}
    \end{figure*}
}

\newcommand{\dynamicboxandsteps}{
    \begin{figure}[t]
      \centering
      \subfloat[Iterative evaluation]
      {
        \includegraphics[width=\linewidth]{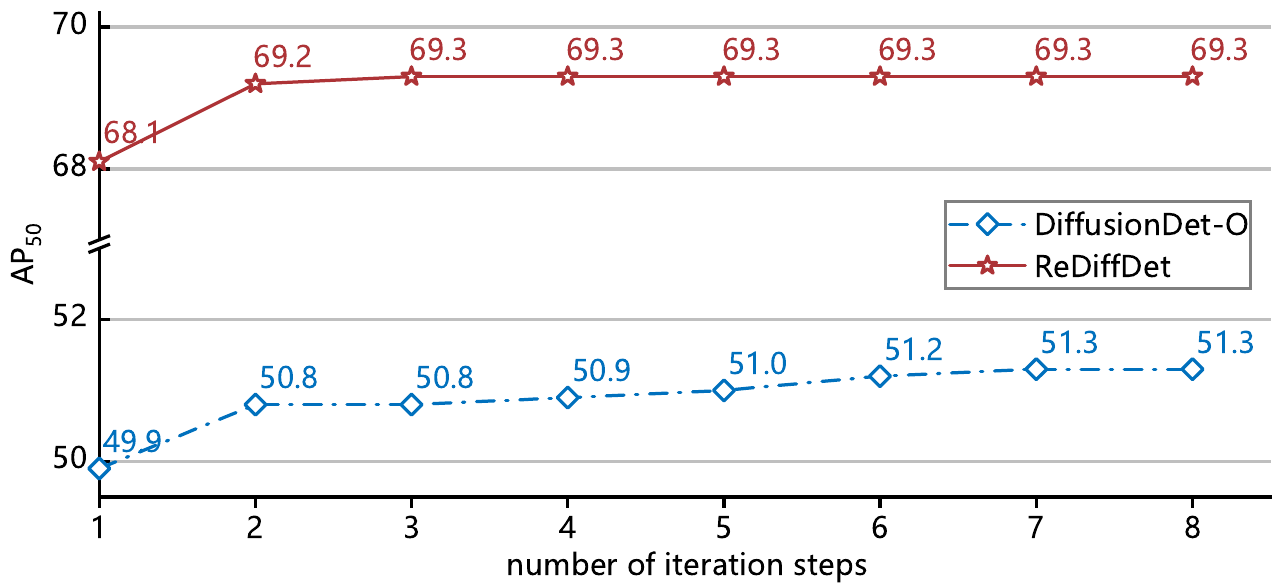}
        \label{ReDiffDetvsDiffusionDetIter}
      }\\
      \subfloat[Dynamic number boxes]
      {
        \includegraphics[width=\linewidth]{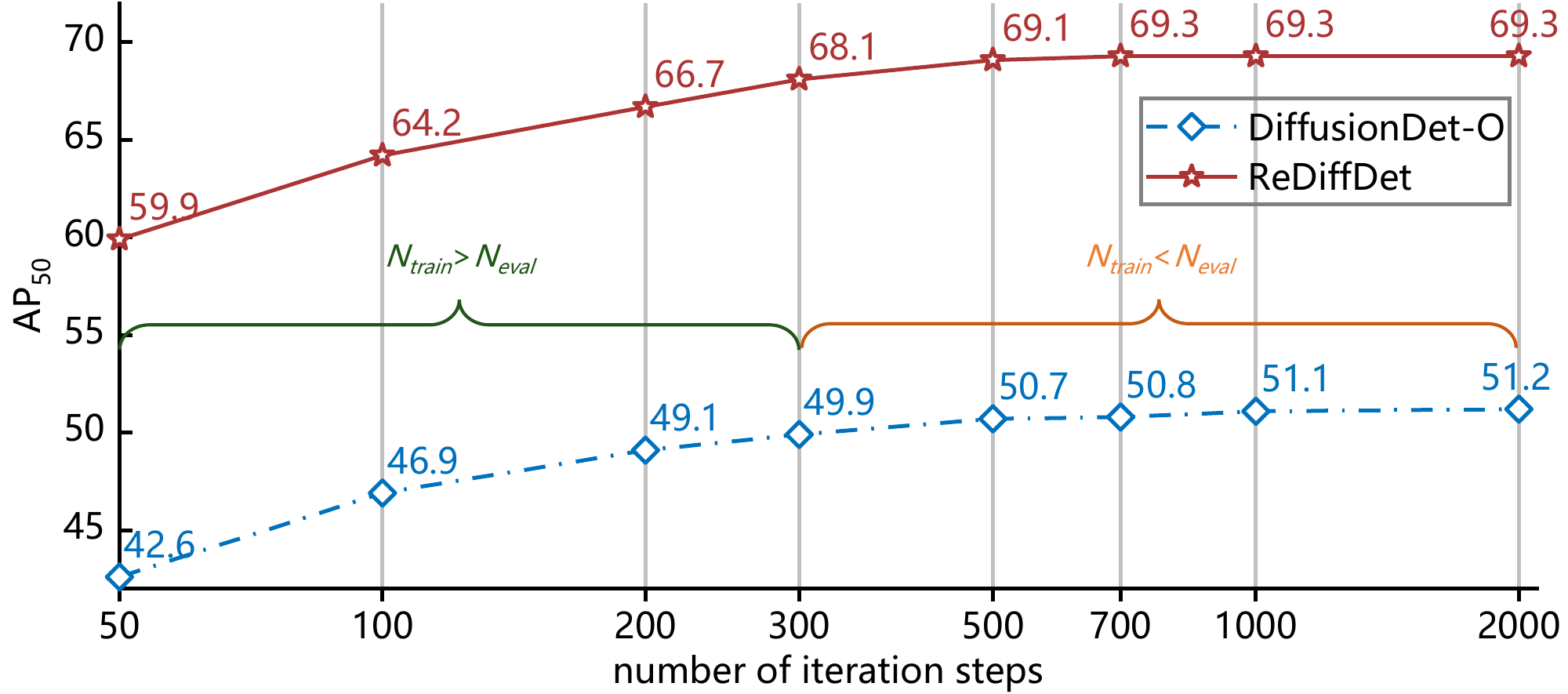}
        \label{ReDiffDetvsDiffusionDetbox}
      }
      \vspace{-0.5em}
       \caption{Results with AP$_{50}$ using \textbf{iterative evaluation} and \textbf{dynamic boxes} on DIOR-R.}
       \label{conditionalencoder}
    \end{figure}

}

\newcommand{\onlyconditiononlynoisemix}{
    \begin{figure}[t]
      \centering
      \includegraphics[width=\linewidth]{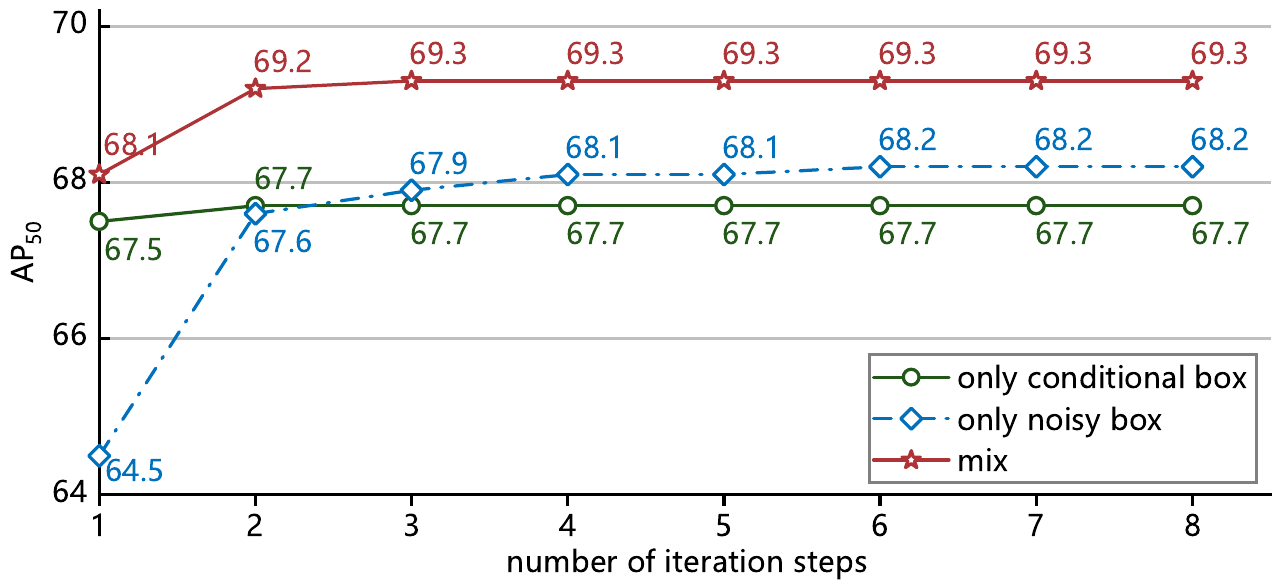}
      \vspace{-2em}
       \caption{Iterative evaluation with input variations: \textbf{only conditional boxes}, \textbf{only noisy boxes}, and \textbf{a combination of both}.}
       \label{onlyconditiononlynoisemix}
    \end{figure}

}

\newcommand{\visualdiffusion}{
    \begin{figure}[t]
      \centering
      \includegraphics[width=\linewidth]{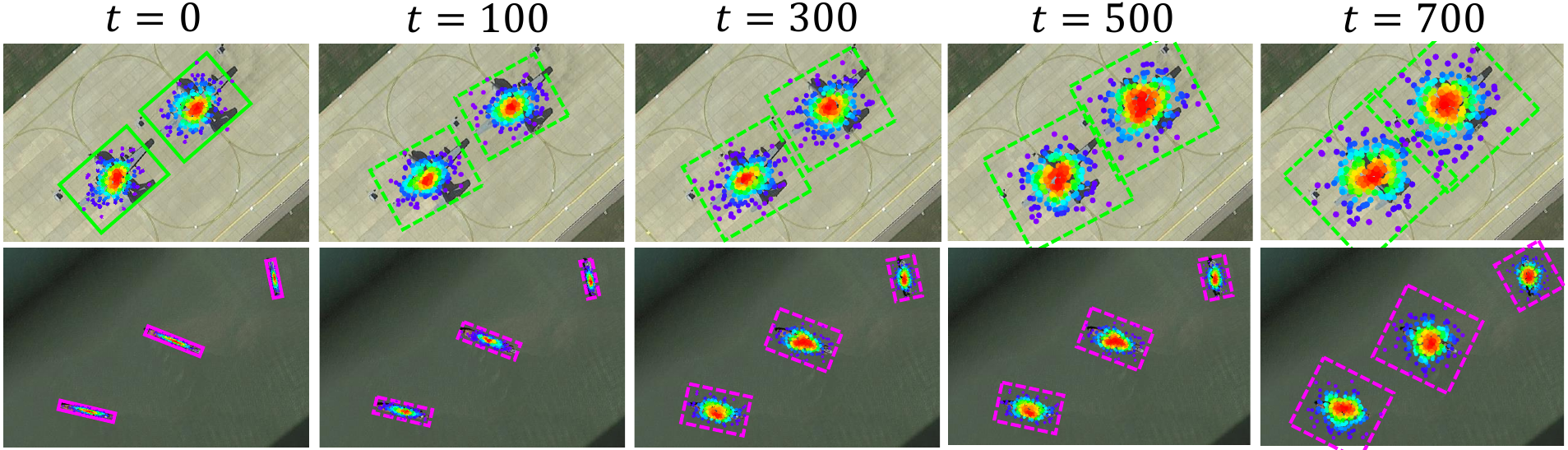}
       \caption{Visualization of \textbf{2D Gaussians for boxes} at different timesteps in diffusion process.}
       \vspace{-0.5em}
       \label{visualdiffusion}
    \end{figure}

}

\newcommand{\visualcenter}{
    \begin{figure}[t]
      \centering
      \includegraphics[width=\linewidth]{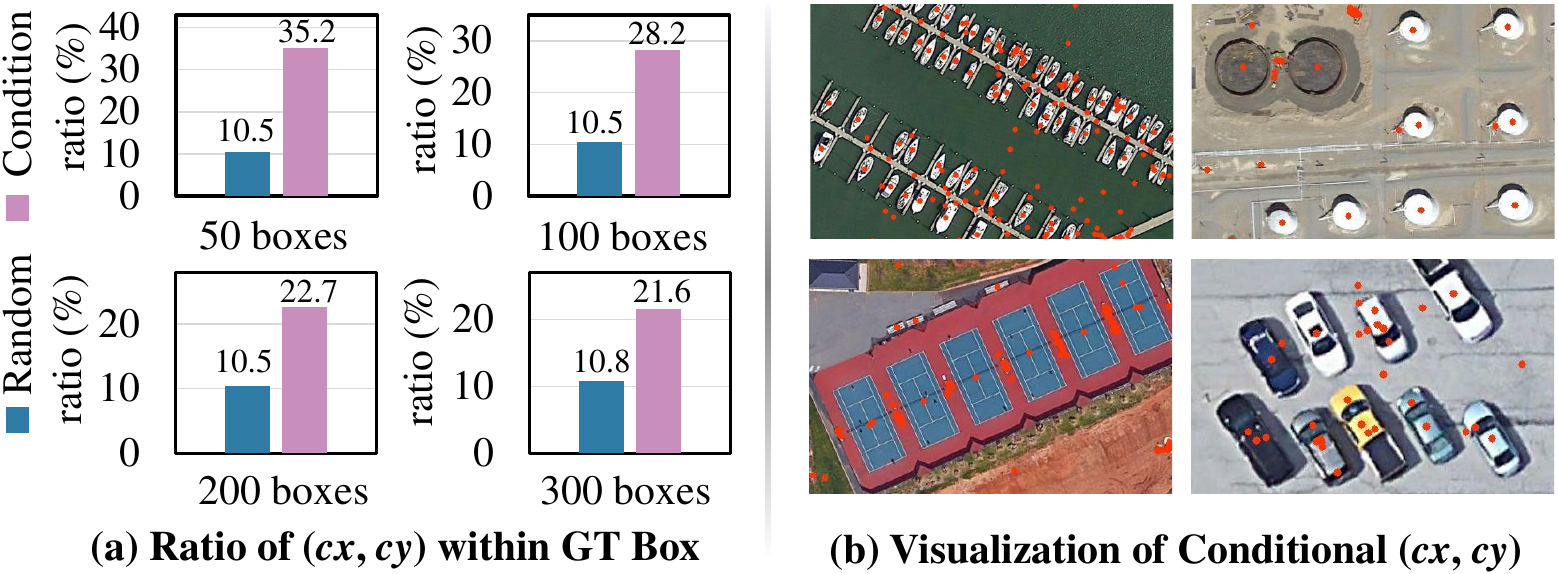}
       \caption{Illustration of centers $(cx,cy)$ of conditional boxes.}
       \vspace{-0.5em}
       \label{visualcenter}
    \end{figure}

}

\newcommand{\visualresult}{
    \begin{figure*}[t]
      \centering
      \includegraphics[width=\textwidth]{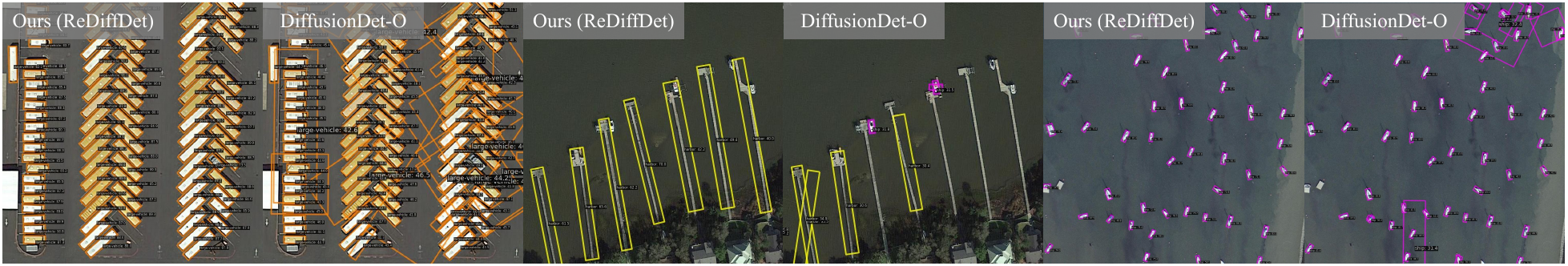}
       \caption{Visualization of \textbf{detection results} for objects with \textbf{dense arrangements}, \textbf{large aspect ratios}, and \textbf{small sizes} on DOTA, such as large vehicles, harbors, and ships.}
       \label{visualresults}
       \vspace{-1.3em}
    \end{figure*}

}

\newcommand{\distributionbox}{
    \begin{figure}[h]
    \centering
    \subfloat[\small][$m$=4]
        {
            \includegraphics[width=0.48\linewidth]{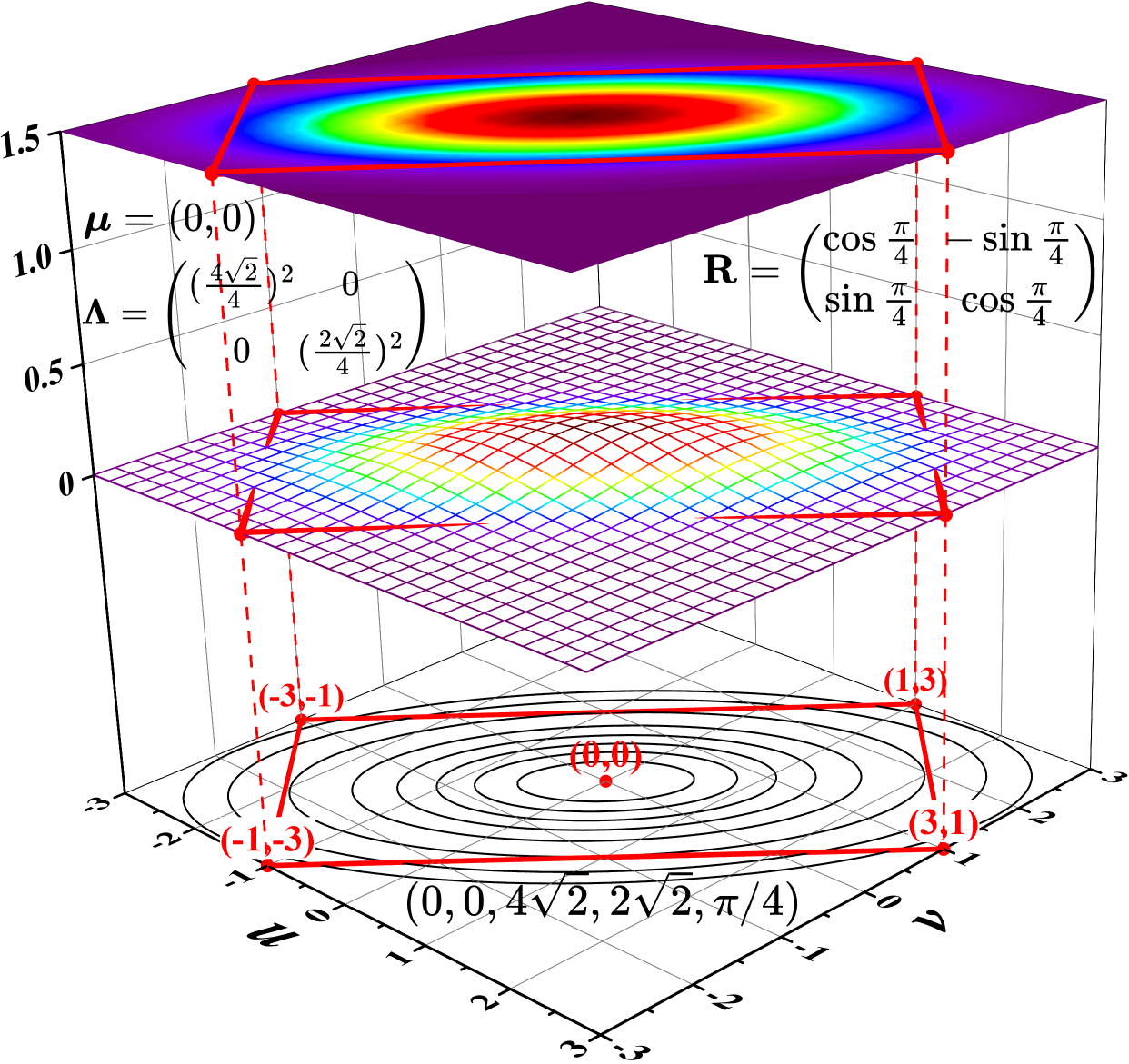}
            \label{lambda_4}
        }
    \subfloat[\small][$m$=6]
        {
            \includegraphics[width=0.48\linewidth]{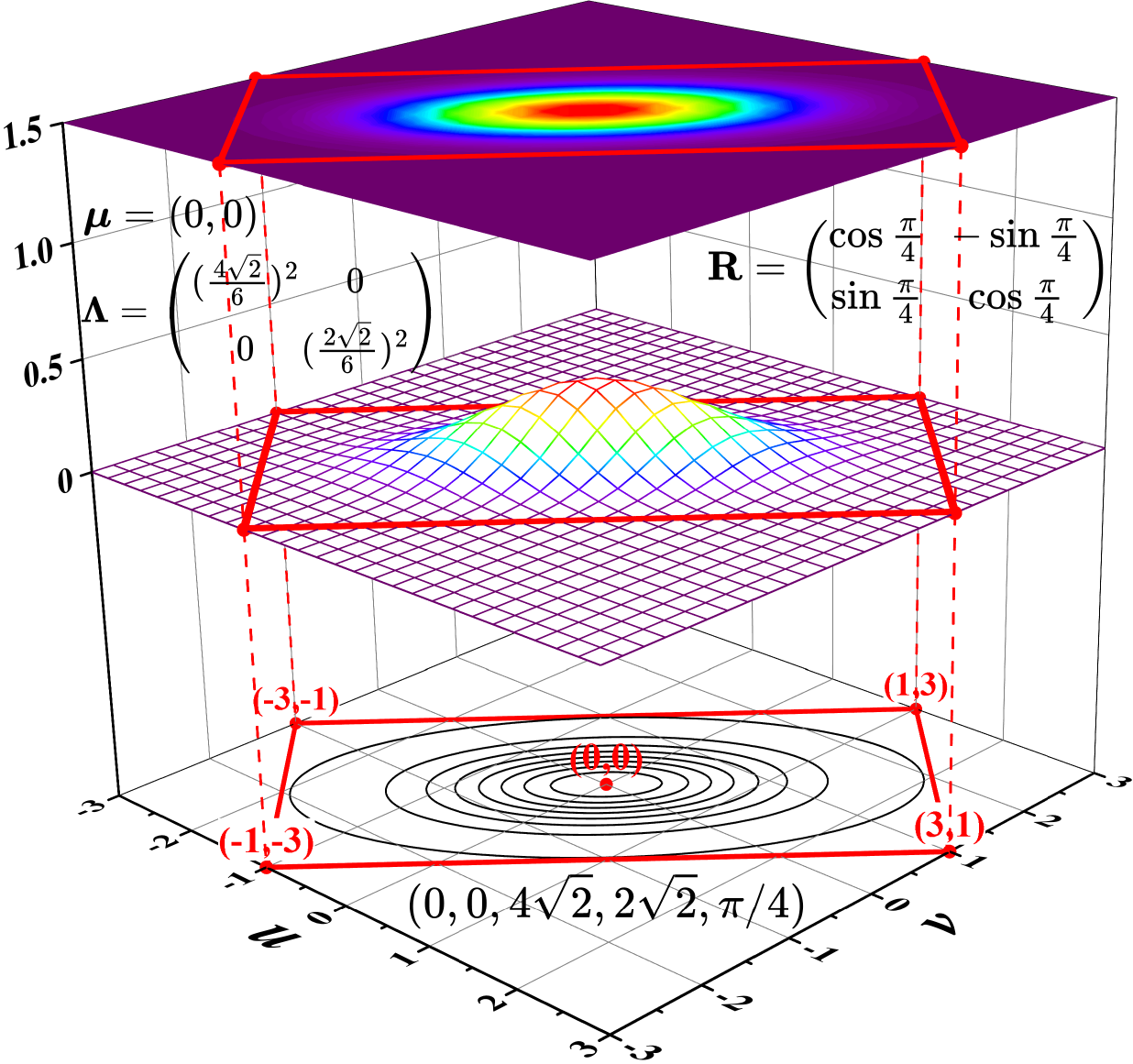}
            \label{lambda_6}
        }
    \caption{Illustration of converting an oriented box to a Gaussian distribution under $m$=4 and $m$=6. Here the example oriented box is $(cx,cy,w,h,a)=(0, 0, 4\sqrt{2},2\sqrt{2},\frac{\pi}{4})$.}
    \label{distribution_box_lambda}
    \end{figure}

}

\begin{document}
\maketitle
\begin{abstract}
The diffusion model has been successfully applied to various detection tasks. However, it still faces several challenges when used for oriented object detection: objects that are arbitrarily rotated require the diffusion model to encode their orientation information; uncontrollable random boxes inaccurately locate objects with dense arrangements and extreme aspect ratios; oriented boxes result in the misalignment between them and image features. To overcome these limitations, we propose ReDiffDet, a framework that formulates oriented object detection as a rotation-equivariant denoising diffusion process. First, we represent an oriented box as a 2D Gaussian distribution, forming the basis of the denoising paradigm. The reverse process can be proven to be rotation-equivariant within this representation and model framework. Second, we design a conditional encoder with conditional boxes to prevent boxes from being randomly placed across the entire image. Third, we propose an aligned decoder for alignment between oriented boxes and image features. The extensive experiments demonstrate ReDiffDet achieves promising performance and significantly outperforms the diffusion model baseline. Codes are available at \url {https://github.com/wokaikaixinxin/ReDiffDet}. \vspace{-1em}
\end{abstract}     
\section{Introduction}
\label{sec:intro}

\teaser

Oriented object detection aims to locate targeted objects by a set of oriented boxes and categorize them~\cite{dotav1.0}. Oriented objects are usually exhibited in arbitrary directions, densely arranged, and with extreme aspect ratios, which make it difficult to accurately predict locations and categories. To predict objects accurately, oriented object detectors based on convolutional neural networks (CNN) have emerged and made considerable progress, mainly including two-stage~\cite{redet,orientedrcnn} and one-stage~\cite{dcfl,fred} method. These methods rely on empirically designed candidates, such as region proposals~\cite{orientedrcnn,redet}, anchor boxes~\cite{dcfl,sasm}, and reference points~\cite{oriented_reppoints,fred}. End-to-end transformers~\cite{arsdetr,emo2detr,orientedformer} have attracted great attention by using a fixed set of learnable queries to eliminate the hand-designed components~\cite{detr}. 

To void the surrogate of queries, DiffusionDet~\cite{diffusiondet} first adopts a diffusion model~\cite{ddim} for horizontal object detection~\cite{coco} and directly detects objects from a set of random boxes. It treats detection as a generative task~\cite{ddpm} over the space of sizes and positions of boxes. In training, DiffusionDet~\cite{diffusiondet} adds noise to ground truth boxes to obtain noisy boxes and learns to predict boxes without noise. In inference, it denoises random boxes to get targets. We further analyze the potential limitations of diffusion-based detectors through two sets of studies in the following sections.

\analysisinintro

We analyzed the distribution of box characteristics, including center coordinates, widths, heights, angles, and aspect ratios, as shown in Fig.~\ref{analysisinintro}. We observe that fitting the distribution of positions and sizes with Gaussian distributions may not be optimal. Only the distribution of center coordinates is more bell-shaped. Different datasets exhibit significant variations in data characteristics. DiffusionDet~\cite{diffusiondet} considers the distribution of coordinates, widths, and heights separately, as multiple 1D distribution. But oriented boxes are geometrically 2D oriented rectangles in a 2D plane. In addition, we conducted a small analytical experiment with the condition that the center coordinates of random boxes be confined to real boxes at the start of inference, as shown in Fig.~\ref{analysisinintro}. The accuracy is improved with the condition, particularly when using a small number of boxes and steps. We found that uncontrollable boxes introduce randomness, but conditioning can help mitigate its effects. This condition prevents the boxes from being randomly located throughout the entire image.

According to the above observations, directly extending the diffusion model into oriented object detection still suffers from the following issues. 1) Equivariance: Objects distributed with arbitrary orientations require the detector to encode orientation information effectively. Ideally, the detector should be able to consistently predict regardless of the orientation of objects. 2) Conditioning: Random boxes do not effectively cover arbitrarily oriented objects or match the real size of objects with extreme aspect ratios and scales. 3) Misalignment: Objects rotate arbitrarily, while image feature maps are axis aligned. Complex background information will be extracted without alignment and key semantic information will be lost.

In this paper, to alleviate the above issues, we propose a rotation-equivariant diffusion model for oriented object detection called ReDiffDet. ReDiffDet casts oriented object detection as a generative task over the space of the Gaussian distribution of boxes. In the training stage, standard Gaussian noise is added to the distribution of ground truth boxes. ReDiffDet is trained to predict the distribution of target boxes from standard Gaussian distribution. In inference, ReDiffDet generates oriented boxes using the reverse process. We theoretically prove the rotation-equivariance in the reverse process under the Gaussian representation and model framework. In addition, we propose a conditional encoder to obtain conditional boxes for conditioning. Finally, we design an aligned decoder to align oriented boxes and image features. In general, the main contributions of this paper are summarized in four folds:
\begin{itemize}
\item As far as we know, this paper proposes the first rotation-equivariant diffusion model for oriented object detection called ReDiffDet for the orientation of boxes.
\item The conditional encoder is designed to produce conditional boxes, helping to prevent the random distribution of boxes across images.
\item The aligned decoder is designed to align oriented boxes with image features by wrapping interesting region features according to the specified boxes.
\item Experiments on five common datasets demonstrate that ReDiffDet is competitive with previously well-established detectors.
\end{itemize} 
\section{Related Work}
\label{sec:relatedwork}

\textbf{Oriented object detection.} CNN methods have achieved great success. One line of work aims to construct dedicated detectors, including backbone~\cite{lsknet, arc}, neck~\cite{r3det}, detection head~\cite{s2anet}, region proposal network~\cite{roi_transformer,orientedrcnn}, rotated region of interest (RRoI) extraction~\cite{orientedrcnn,redet} and label assignment~\cite{sasm}. Another line of work focuses on more effective representations of angles~\cite{xu2024rethinking,xiao2024theoretically}. The third line of work design loss function~\cite{gwd, kld} for better supervision. The fourth line of work focuses on aligning oriented boxes with image features, such as modified convolution~\cite{s2anet, arc} and RRoI extraction~\cite{orientedrcnn,redet}. End-to-end transformers demonstrate competitive performance without hand-crafted post processing components~\cite{detr}. Modern end-to-end methods have improved encoder~\cite{ao2detr}, query~\cite{orienteddetr, d2qdetr, psd-sq}, attention~\cite{orientedformer, arsdetr, rqformer}, and label assignment~\cite{emo2detr} for oriented positional queries respectively.

Diffusion models for oriented object detection have not been fully explored. DiffusionDet~\cite{diffusiondet} is specifically designed for horizontal object detection. The differences between our ReDiffDet and DiffusionDet~\cite{diffusiondet} are as follows: (1) ReDiffDet is designed for oriented boxes, while DiffusionDet is for horizontal boxes; (2) ReDiffDet converts a box to a 2D Gaussian distribution and establishes a diffusion process based on it, while DiffsuionDet models positions and sizes as multiple 1D Gaussian distributions; (3) ReDiffDet builds a rotation-equivariant diffusion model, while DiffusionDet does not; (4) ReDiffDet introduces a conditioning strategy to restrict boxes, while DiffusionDet adopts a set of random boxes; (5) ReDiffDet aligns oriented boxes with image features, while DiffusionDet does not.

\textbf{Rotation-equivariance.} This property means that applying rotation transformations to the input results in predictable rotations of the feature. ReDet~\cite{redet} designs a first rotation-equivariant detector for oriented object detection. It reconstructs the backbone and feature pyramid network using steerable group CNNs~\cite{g_cnns} and introduces RiRoI Align to generate rotation-invariant features. FRED~\cite{fred} further decouples the invariant classification task and the equivariant task to achieve fully rotation-equivariant oriented object detection. Unlike these CNN methods, ReDiffDet is based on a diffusion model.

\textbf{Conditioning in diffusion model.} Conditions allow the diffusion model to generate content of a specific style based on given conditions~\cite{stablediff,instancediffusion}, such as texts~\cite{controlnet,kondapaneni2024text} and semantic information~\cite{semanticonditiondiff,matfuse}. Conditions help control the generated content, improve sample quality, reduce uncertainty in the generation process, process multimodal data, and maintain data structure integrity~\cite{stablediff}. Different from existing studies, we propose a conditional encoder to learn conditional boxes, which can provide object location priors, effectively avoiding a large number of random initial boxes and reducing the uncertainty of object positioning.

\section{Methods}
\label{sec:method}

\overview

\subsection{Preliminaries}
\label{Sec_equivariance}
\textbf{2D Rotation Group.}
We primarily focus on the 2D rotation group $SO(2)$ to describe the orientation in space $\mathbb{R}^2$, \ie, 2D image plane. It is a continuous group where each element is associated with a rotation about the origin by an angle $a$. The elements are represented by rotation matrices $\mathbf{R}=[[\cos a,\sin a]^\top,[-\sin a,\cos a]^\top]$, which are $2\times2$ orthogonal matrices with $\mathrm{det}(\mathbf{R})=1$.

\textbf{Equivariance and Invariance.}
Given a transformation group $G$, a function $\rho:X\to Y$ is said to be equivariant if:
\begin{equation}\label{ori_equivariance}
    \rho(S^X_g(x))=S^Y_g(\rho(x)),
\end{equation}
where $g\in G$, $x\in X$. The $S^{(X)}_g$ and $S^{(Y)}_g$ denote group actions in the space $X$ and $Y$ respectively. Specially, the invariance of the function $\rho(\cdot)$ can be expressed as:
\begin{equation}\label{ori_invariance}
  \rho(S^X_g(x))=\rho(x).
\end{equation}
In this work, the group $G$ refers to $SO(2)$. The function $\rho(\cdot)$ refers to the proposed network, which takes in distribution and outputs new distribution. The group action $S^{(\cdot)}_g$ rotates sample data around the origin.

\subsection{Box-oriented Diffusion Model}
\textbf{Gaussian Modeling.} We represent an oriented box as a 2D Gaussian, serving as the initial distribution for the diffusion process, as illustrated in Fig.~\ref{overview}. The transformation between oriented boxes and distributions is introduced below.

1) \textit{Oriented box$\to$2D Gaussian.} An oriented box $(cx, cy, w, h, a)$ representing center coordinates, widths, heights, and angles, can be converted into a 2D Gaussian distribution $\mathcal{N}(\boldsymbol{\mu},\space \boldsymbol{\Sigma})$ as follow:
\begin{equation}
    \begin{split}
            \boldsymbol{\mu}&=(cx, cy)^{\top},\quad\mathbf{\Sigma}=\mathbf{R}\mathbf{\Lambda}\mathbf{R}^{\top}, \\
            \mathbf{R}&=\begin{pmatrix}
                  \cos a &  -\sin a  \\
                  \sin a &  \cos a
            \end{pmatrix},\quad\mathbf{\Lambda}=
            \begin{pmatrix}
                  (\frac{w}{m})^2&  0 \\
                  0&  (\frac{h}{m})^2
            \end{pmatrix},
    \end{split}
    \label{box_to_gau}
\end{equation}
where the expectation is denoted by $\boldsymbol{\mu}$, the variance by $\mathbf{\Sigma}$, the rotation matrix by $\mathbf{R}$, and the diagonal matrix of eigenvalues by $\mathbf{\Lambda}$. The scaling factor $m$ adjusts distributions according to the 68-95-99.7 rule~\cite{rohatgi2015introduction}. Furthermore, the probability density function (PDF) is calculated as follows:
{\small
\begin{equation}
\mathrm{PDF}(u,v)=\frac{\exp (-\frac{1}{2} ((u,v)^{\top}\hspace{-1.5mm}-\boldsymbol{\mu})^\top  \boldsymbol{\Sigma}^{-1}((u,v)^{\top}\hspace{-1.5mm}-\boldsymbol{\mu}))}{2\pi \sqrt{\left | \boldsymbol{\Sigma}  \right | } }. 
\end{equation}
}

2) \textit{2D Gaussian$\to$sample.} Given the mean $\boldsymbol{\mu}$ and variance $\mathbf{\Sigma}$, we sample $k$ data points $P=\{\mathbf{P}_1,...,\mathbf{P}_k\}=\{(u_1,v_1),...,(u_k,v_k)\}$. Samples that follow $\mathcal{N}(\boldsymbol{\mu},\space \boldsymbol{\Sigma})$ can be obtained by successively applying scaling, rotation, and translation to samples from $\mathcal{N}(\mathbf{0},\mathbf{I})$ using the diagonal matrix $\boldsymbol{\Lambda}$, the rotation matrix $\mathbf{R}$, and the coordinate $(cx,cy)$.

3) \textit{Sample$\to$2D Gaussian.} Given samples $\{\mathbf{P}_1,$ $\mathbf{P}_2,$ $...,$ $\mathbf{P}_k\}$, the mean $\boldsymbol{\mu}$ and variance $\boldsymbol{\Sigma}$ are estimated using the maximum likelihood estimation (MLE):
\begin{equation}
    \hat{\boldsymbol{\mu}},\hat{\boldsymbol{\Sigma}}=\arg \max_{\boldsymbol{\mu},\boldsymbol{\Sigma}} \prod_{i=1}^{k}\mathrm{PDF}(\mathbf{P}_i|\boldsymbol{\mu},\boldsymbol{\Sigma})
  \label{sample_to_gau},
\end{equation}
where $\hat{\boldsymbol{\mu}}$ and $\hat{\boldsymbol{\Sigma}}$ denote estimators.

4) \textit{2D Gaussian$\to$oriented box.} Given the mean $\boldsymbol{\mu}$, the center coordinates $(cx,cy)$ are calculated as:
\begin{equation}
  (cx,\, cy)^\top=\boldsymbol{\mu}.
\end{equation}
Given variance $\boldsymbol{\Sigma}$, the matrix $\mathbf{R}$ and $\mathbf{\Lambda}=\mathrm{diag}(\lambda_1,\lambda_2)$ are calculated using the eigenvalue decomposition (EVD):
\begin{equation}
  \det(\boldsymbol{\Sigma}-\lambda\mathbf{I})=0, \quad (\boldsymbol{\Sigma}-\lambda_i\mathbf{I})\mathbf{R}_{:,i}=\mathbf{0},(i=1,2),
  \label{gau_to_box}
\end{equation}
where $\lambda$ represents the eigenvalues, specifically $\lambda_1,\lambda_2$. The notation $\mathbf{R}_{:,i}$ refers to the $i$-th column vector of the matrix $\mathbf{R}$. The size and angle can be further calculated~\cite{mmrotate}.

\textbf{Diffusion Process.} The data $\bm{z}_0$ is sampled from a 2D Gaussian distribution $\mathcal{N}(\boldsymbol{\mu},\boldsymbol{\Sigma})$, converted from corresponding oriented box $(cx,cy,w,h,a)$. The origin of the coordinate system is placed at the box center coordinate. We define a Markovian chain by gradually adding standard Gaussian noise to the data at each timestep $t\in\{0,1,2...,T\}$, starting from the initial data $\bm{z}_0\sim q(\bm{z}_0)$. The diffusion process is defined as:
\begin{equation}
  q(\bm{z}_{t}|\bm{z}_0)=\mathcal{N}(\bm{z}_t;\sqrt{\bar{\alpha}_t}\bm{z}_0,(1-\bar{\alpha}_t)\mathbf{I}),
\end{equation}
where $\alpha_t=1-\beta_t$ and $\bar{\alpha}_t= {\textstyle \prod_{i=1}^{t}}\alpha_i$. The $\beta_t\in(0,1)$ denotes the variance schedule~\cite{ddpm}. When $t\to T$, the $q(\bm{z}_t)$ approaches an isotropic Gaussian distribution, \ie, $\mathcal{N}(\mathbf{0}, \mathbf{I})$. At the training stage, the network $\epsilon_\theta(\bm{z}_{t},t)$ is trained to predict $\bm{z}_0$ from $\bm{z}_t$ by minimizing the $\ell_2$ loss~\cite{ddpm,diffusiondet}.

\textbf{Reverse Process.} The goal of the reverse process $p_{\theta}(\cdot)$ is learning to recover target Gaussian distribution $\bm{z}_0$ from the noise $\bm{z}_T$. Next, we provide a theoretical analysis of the equivariance and invariance for the reverse process.

1) Time \textit{t=T}. Conditional boxes are provided by the conditional encoder at time $T$, whose center coordinates are used as origins for corresponding distributions. At the beginning of the reverse process, noises are sampled from standard Gaussian distribution, \ie, $\bm{z}_T\sim\mathcal{N}(\mathbf{0},\mathbf{I})$. Formally, let $\mathbf{R}\in SO(2)$ be the rotation, we have:

\noindent
\textbf{\textit{Proposition 1.}} \textit{At time $T$ in the reverse process $p_{\theta}(\cdot)$, the function $p(\bm{z}_T)$ is rotation invariant, i.e., $p(\bm{z}_T) = p(\mathbf{R}(\bm{z}_T))$ under $SO(2)$. Proof:}

\noindent The mean and variance of $\bm{z}_T$ are calculated by Eq. (\ref{mean_var_z_T}).
\begin{equation}
  \mathbb{E}(\bm{z}_T)=\mathbf{0}, \; \mathbb{V}(\bm{z}_T)=\mathbf{I}.
  \label{mean_var_z_T}
\end{equation}
The mean and variance of {\small $\mathbf{R}(\bm{z}_T)$} are calculated by Eq. (\ref{mean_var_R_z_T}).
\begin{equation}
\begin{split}
  &\mathbb{E}(\mathbf{R}(\bm{z}_T))=\mathbf{R}\mathbb{E}(\bm{z}_T)=\mathbf{0},\\
  &\mathbb{V}(\mathbf{R}(\bm{z}_T))=\mathbf{R}\mathbb{V}(\bm{z}_T)\mathbf{R}^\top=\mathbf{I},
\end{split}
\label{mean_var_R_z_T}
\end{equation}
where $\mathbf{R}\mathbf{R}^T=\mathbf{I}$. It is proved that $\mathbb{E}(\bm{z}_T)=\mathbb{E}(\mathbf{R}(\bm{z}_T))$ and $\mathbb{V}(\bm{z}_T)=\mathbb{V}(\mathbf{R}(\bm{z}_T))$. This indicates that the isotropic Gaussian $p(\bm{z}_T)$ is invariant to rotations around the origin.

2) Time \textit{1$<$t$<$T}. Inspired by~\cite{diffusiondet}, DDIM sampling strategy~\cite{ddim} is adopted, which allows for much faster sampling. When $1<t<T$, the reverse process is defined as:
\begin{equation}
  p_\theta(\bm{z}_{t-1}|\bm{z}_t)=\mathcal{N}(\bm{z}_{t-1};\boldsymbol{\mu}_{\theta}(\bm{z}_t,t),\boldsymbol{\Sigma}_{\theta}(\bm{z}_t,t)).
\end{equation}
Formally, let $\mathbf{R}\in SO(2)$ be the rotation, we have:

\noindent
\textbf{\textit{Proposition 2.}} \textit{During $1<t<T$ in the reverse process $p_\theta$, the function $p_\theta(\bm{z}_{t-1}|\bm{z}_t)$ is rotation equivariant, i.e., $p_\theta(\bm{z}_{t-1}|\bm{z}_t)=p_\theta(\mathbf{R}(\bm{z}_{t-1})|\mathbf{R}(\bm{z}_t))$ under $SO(2)$, as long as $\mathbf{R}(\boldsymbol{\mu}_{\theta}(\bm{z}_t,t))=\boldsymbol{\mu}_{\theta}(\mathbf{R}(\bm{z}_t),t)$. Proof:}
\begin{align}
    &p_\theta(\mathbf{R}(\bm{z}_{t-1})|\mathbf{R}(\bm{z}_t))\nonumber\\
    &= \frac{1}{2\pi\sqrt{\left | \boldsymbol{\Sigma}_\theta(\mathbf{R}(\bm{z}_t),t)\right|}} \mathrm{exp} \{-\frac{1}{2}(\mathbf{R}(\bm{z}_{t-1})-\boldsymbol{\mu}_\theta(\mathbf{R}(\bm{z}_t),t)) ^\top\nonumber\\  
    &\quad\cdot\boldsymbol{\Sigma}_\theta(\mathbf{R}(\bm{z}_t),t)^{-1}\cdot(\mathbf{R}(\bm{z}_{t-1})-\boldsymbol{\mu}_\theta(\mathbf{R}(\bm{z}_t),t))\}\nonumber\\
    &= (2\pi\sqrt{\left |\sigma ^2_t\mathbf{I} \right|})^{-1} \mathrm{exp} \{-\frac{1}{2}(\bm{z}_{t-1}-\boldsymbol{\mu}_\theta(\bm{z}_t,t))^\top\nonumber\\  
    &\quad\cdot (1/\sigma^2_t)\mathbf{I} \cdot(\bm{z}_{t-1}-\boldsymbol{\mu}_\theta(\bm{z}_t,t))\}=p_\theta(\bm{z}_{t-1}|\bm{z}_t).
\end{align}
Then, the key for Proposition 2 is to make sure the mean $\boldsymbol{\mu}_\theta(\bm{z}_t,t)$ satisfies $\mathbf{R}(\boldsymbol{\mu}_{\theta}(\bm{z}_t,t))=\boldsymbol{\mu}_{\theta}(\mathbf{R}(\bm{z}_t),t)$ for rotation equivariance. The $\boldsymbol{\mu}_\theta(\bm{z}_t,t)$ is one of outputs of the neural network $\epsilon_\theta(\bm{z}_t,t)$. The problem now becomes ensuring that the model $\epsilon_\theta(\bm{z}_t,t)$ maintains rotation equivariance. We borrow the ReResNet with ReFPN and RiRoI Align from ReDet~\cite{redet} to extract rotation-equivariant and invariant features. 

3) Time \textit{t=1}. The $p(\bm{z}_0)$ can be derived from the previous state. Formally, let $\mathbf{R}\in SO(2)$ be the rotation, we have:

\noindent
\textbf{\textit{Proposition 3.}} \textit{At time $t=1$ in the reverse process $p_\theta$, given the rotation invariant $p(\bm{z}_T)$, i.e., $p(\bm{z}_T)=p(\mathbf{R}(\bm{z}_T))$, and the rotation equivariant $p_\theta(\bm{z}_{t-1}|\bm{z}_t)$, i.e., $p_\theta(\bm{z}_{t-1}|\bm{z}_t)=p_\theta(\mathbf{R}(\bm{z}_{t-1})|\mathbf{R}(\bm{z}_t))$, then the $p_\theta(\bm{z}_0)$ is also rotation invariant, i.e., $p_\theta(\bm{z}_0)=p_\theta(\mathbf{R}(\bm{z}_0))$ under $SO(2)$. Proof:}
\vspace{-0.3em}
\begin{equation}
\begin{split}
    &p_\theta(\mathbf{R}(\bm{z}_0))=\hspace{-0.5mm}\int\hspace{-0.5mm} p(\mathbf{R}(\bm{z}_T)) {\textstyle \prod_{t=1}^{T}} p_\theta(\mathbf{R}(\bm{z}_{t-1})|\mathbf{R}(\bm{z}_t))\mathrm{d} \bm{z}_{1:T}\\
    &=\hspace{-0.5mm}\int \hspace{-0.5mm} p(\bm{z}_T){\textstyle \prod_{t=1}^{T}} p_\theta(\bm{z}_{t-1}|\bm{z}_t)\mathrm{d} \bm{z}_{1:T}=p_\theta(\bm{z}_0).
\end{split}
\end{equation}

\Algorithmtrain

\vspace{-0.6em}

\subsection{Conditional Encoder} 
The rotation-equivariant features extracted from raw images through the backbone with feature pyramid network~\cite{fpn} are fed into the conditional encoder as inputs. This encoder consists of two main components, a box subnet and a score subnet, both utilizing simple convolutional layers to generate boxes and scores, respectively~\cite{ddq}. To guide the aligned decoder in denoising, we select the top boxes based on their scores as conditional boxes. This design choice is motivated by our belief that a learnable network offers greater effectiveness compared to an unlearnable strategy for controlling boxes. During the training stage, conditional boxes are first transformed into Gaussians, and then the ground truth Gaussians are padded so that all distributions sum to a fixed number, as shown in Algo.~\ref{algorithmtrain}. In the inference procedure, the center coordinates of the conditional boxes are set as the origin of the coordinate system, and the size determines the scale of distribution, as shown in Algo.~\ref{algorithminference}.

\Algorithminference

\subsection{Aligned Decoder} 
The decoder stacks multiple decoder layers, each layer consists of RiRoI Align~\cite{redet}, self-attention, and instance interaction~\cite{diffusiondet} successively. The noise distribution or the estimated distributions from the last sampling step are sent into the decoder. The input distributions are transformed into oriented boxes. For these boxes, RiRoI Align~\cite{redet} warp region features to ensure alignment and rotation-invariance. Next, the self-attention captures pairwise interactions between the RiRoI-features. Then, time embedding encodes the timestep for the diffusion model and interacts with the attention results. The linear operator is appended at the end and outputs the classes and boxes. The outputs can be used for the next layer or iterative evaluation. In training, predictions are supervised by set prediction loss~\cite{diffusiondet}. In inference, we also adopt the box renewal strategy~\cite{diffusiondet} to revive the high-scoring predictions.

\vspace{-0.5em}
\section{Experiment}
\label{sec:experiment}

\DOTAOneResults

\subsection{Datasets and Implementation Details}
\textbf{Datasets.} We conducted experiments on 5 datasets. DIOR-R~\cite{dior} and DOTA-v1.0~\cite{dotav1.0}/1.5/2.0~\cite{dota-v2.0} are large-scale datasets for oriented object detection, using the \texttt{trainval} set for training and \texttt{test} set for testing. DIOR-R~\cite{dior} contains 23,463 images, each sized $800 \times 800$, across 20 classes. DOTA-v1.0~\cite{dotav1.0} contains 2,806 images with 15 categories. DOTA-v1.5~\cite{dotav1.0} uses the same images as DOTA-v1.0 and has 16 categories. DOTA-v2.0~\cite{dota-v2.0} has 18 categories and 11,268 images. COCO~\cite{coco} is a dataset for horizontal object detection, containing 80 categories, with 118K images for training and 5K images for evaluation.

\TableDOTAVonefiveandtwoResult

\textbf{Implementation Details}. We conduct experiments using 2 NVIDIA RTX 2080ti GPUs, with a batch size of 4. The model is optimized using the AdamW~\cite{adamw} with the weight decay $10^{-4}$. The initial learning rate is set to $4 \times 10^{-5}$ for DIOR-R and $5.5 \times 10^{-5}$ for DOTA-v1.0/1.5/2.0. The training schedule consists of 24 epochs, with the learning rate reduced by a factor of 10 at 16 and 22 epochs. Only random flipping is used as the data augmentation strategy. The code is built upon MMRotate~\cite{mmrotate} and PyTorch with random seed. Images in DOTA-v1.0/1.5/2.0 are cropped into $1024 \times 1024$ patches with 200-pixel overlaps for single-scale training. The performance on the DOTA test set is obtained from the official evaluation server. The ReResNet50 with ReFPN~\cite{redet} is the default backbone if not specified. For COCO, the settings are the same as in DiffusionDet~\cite{diffusiondet}, except the batchsize is set to 4 across 2 GPUs, and the training schedule is set to 12 epochs.

\vspace{-0.5em}
\subsection{Main Results}

\textbf{Results on DOTA series.} We compare ReDiffDet with modern one-stage, two-stage, end-to-end, and diffusion detectors. The results for each class on DOTA-v1.0~\cite{dotav1.0} are presented in Tab.~\ref{tab:dota1.0}. We use 300 boxes and 1 iteration for testing by default. ReDiffDet achieves 75.45\% AP$_{50}$ with ReResNet50~\cite{redet} and 80.70\% AP$_{50}$ under multi-scale training, demonstrating its competitiveness across various detection methods. Additionally, ReDiffDet achieves 68.00\% AP$_{50}$ on DOTA-v1.5 and 55.76\% AP$_{50}$ on DOTA-v2.0~\cite{dota-v2.0}, as shown in Tab.~\ref{tab:dota1.5and2.0}. We further evaluate our method using the ResNet50~\cite{resnet} and Swin-T~\cite{swin} backbones. Although these backbones cannot extract rotation-equivariant features, the model still follows the diffusion and reverse process. We suggest that the improvement is due to the larger number of parameters in the backbone.

\DIORRResults

\textbf{Results on DIOR-R.} We report main results of some small size objects on DIOR-R in Tab.~\ref{tab:diorr}. Detailed results per category are provided in the Appendix. Our model achieves 65.47\% AP$_{50}$ with PKINet-T~\cite{pkinet}, 66.62\% AP$_{50}$ with LSK-T~\cite{lsknet}, and 68.05\% AP$_{50}$ with ReResNet50~\cite{redet}, surpassing all compared methods. Our method significantly improves accuracy for objects with small sizes and large aspect ratios compared with DiffusionDet-O~\cite{diffusiondet}.

\vspace{-0.5em}
\COCOResults
\TableAblationStudiesIntegrated
\textbf{Results on COCO.} Although our approach focuses on oriented boxes, the conditioning strategies can be extended to diffusion-based horizontal object detection. We evaluate DiffusionDet's performance on the COCO dataset with and without our proposed conditional encoder, as shown in Tab.~\ref{tab:coco}. Our conditional encoder and boxes achieve a significant improvement of 1.8\% in mAP and substantially enhance performance across various AP metrics.

\dynamicboxandsteps

\vspace{-0.4em}
\subsection{Ablation Study}
\textbf{Decoder layer.} We examine the effect of decoder layers in Tab.~\ref{tab:decoder_layer}. As the number of decoder layers increases, the model's parameters increase. Each additional decoder layer adds approximately 14M parameters. With 4 decoder layers, the model achieves the best performance, with 68.05\% AP$_{50}$ and 49.96\% AP$_{75}$. Therefore, we adopt 4 decoder layers to balance accuracy and model complexity.

\textbf{Scaling factor.} We study the effect of the scaling factor $m$ in Tab.~\ref{tab:scaling_factor}. When the value of $m$ is smaller, the variance of a 2D Gaussian transformed by an oriented box is larger, and the main projection area on the 2D plane increases geometrically. However, if the scaling factor $m$ is too large or too small, the oriented box becomes susceptible to noise. With $m=6$, the model achieves its best performance.

\onlyconditiononlynoisemix 

\textbf{Samples.} We compare the effect of different number $k$ of samples in Tab.~\ref{tab:samples}. As the number of samples increases, the accuracy decreases slightly. The smaller the number of samples, the change in sample expectation and variance is more obviously affected by noise. We adopt $k=8$ for the number of samples.

\textbf{Signal scale.} The results in Tab.~\ref{tab:signal_scale} demonstrate the effect of the signal scale. In our method, noise is added to the sampled points of the boxes. Geometrically, the samples are scaled within the range $[-2, 2]$ when the signal scale is 2.0, which closely aligns with the main range $[-3, 3]$ of the normal noise. However, if the signal scale is too large or too small, the distribution of the sampled data is affected, causing the noisy boxes to deviate from the real objects.

\textbf{Individual strategy.} We check the effectiveness of each proposed strategy in the proposed method. As seen in Tab.~\ref{tab:individual}, the baseline DiffusionDet yields a result of 49.91\% AP$_{50}$ on DIOR-R. When we gradually apply the rotation-equivariant diffusion model, conditional encoder, and aligned decoder into the detector, the accuracy improves progressively, verifying each design's effectiveness.

\visualresult

\vspace{-0.4em}
\subsection{Analysis}
\textbf{Iterative evaluation.} We investigate the performance by increasing the number of iterative steps from 1 to 8, as shown in Fig.~\ref{ReDiffDetvsDiffusionDetIter}. We use ensemble~\cite{diffusiondet} of the output of multiple steps as the final results after non-maximum suppression. The accuracy of both our method and DiffusionDet-O~\cite{diffusiondet} increases as the number of iterative steps grows. However, our ReDiffDet achieves peak accuracy with relatively few iterations. Furthermore, our method can update three different types of input boxes, as shown in Fig.~\ref{onlyconditiononlynoisemix}. For noisy boxes, the accuracy gradually increases to its peak over multiple iterations. For conditional boxes, the model achieves higher accuracy in the first iteration due to the more accurate box shape. However, accuracy does not significantly improve with further iterations, as the box distribution differs from the training data. For mixed boxes, the conditional boxes enhance the denoising process.

\visualdiffusion
\visualcenter

\textbf{Dynamic number of boxes.} We compare the dynamic number of box property of ReDiffDet with DiffusionDet-O~\cite{diffusiondet} on DIOR-R, as shown in Fig.~\ref{ReDiffDetvsDiffusionDetbox}. Both of them are trained with 300 boxes. The evaluation is on $\{50, 100, 200, 300, 500, 700, 1000, 2000\}$ boxes. The accuracy of all methods increases as the number of boxes increases when $N_{eval}<N_{train}$. In contrast to DiffusionDet-O~\cite{diffusiondet}, ReDiffDet can peak without relying on a large number of extra boxes when $N_{eval}>N_{train}$.

\textbf{Results visualization.} We visualize the predicted results in Fig.~\ref{visualresults}. Our method achieves better results than the baseline on objects with small sizes, large aspect ratios, and dense arrangements. In DiffusionDet-O~\cite{diffusiondet}, certain boxes are incorrectly positioned on the background area.

\TableParamsandFPS

\textbf{Geometric perspective on diffusion model.} We visualize the 2D Gaussian distributions of boxes at different timesteps $t$ on the image plane, offering a new geometric perspective, as shown in Fig.~\ref{visualdiffusion}. At $t=0$, the 2D Gaussians are derived directly from oriented boxes. As $t$ increases, the proportion of noise grows, gradually transforming the distribution of boxes into a standard normal distribution.

\textbf{Conditioning strategy.} We calculate the ratios of center coordinates $(cx,cy)$ for both random and conditional boxes that fall within ground truth boxes on the DIOR-R \texttt{test} set, as shown in Fig.~\ref{visualcenter}. The results show that the ratio for conditional boxes is significantly higher than that for random boxes. Furthermore, we visualize the conditional center coordinates, revealing that they are more concentrated around potential foreground objects.

\textbf{Parameters and speed.} The comparison of parameters and inference speed are shown in Tab.~\ref{tab:paramsandfps}. Our method reduces the parameters with ReResNet-50~\cite{redet} compared to the baseline. Additionally, the inference speed is faster than that of compared transformer-based methods.

\vspace{-0.5em}
\section{Conclusion}
In this work, we are the first to explore the rotation-equivariant diffusion model for oriented object detection by formulating the detection as a denoising diffusion process, transforming noise into a distribution of boxes. ReDiffDet has several key properties, including equivariance, conditioning, and alignment. Extensive experiments demonstrate that our method outperforms the baseline and achieves favorable performance compared to traditional detectors. Although our method demonstrates the rotation-equivariance of the reverse process, the properties of the other sampling strategy warrant further discussion. Various geometric shapes, such as 3D boxes, ellipses, and spherical boxes, can be converted into Gaussian or other distributions. They can also be modeled as diffusion processes using our approach.

\clearpage
\section{Acknowledgement}
This work was supported in part by the National Natural Science Foundation of China under Grant 62272461, Grant 62172417, and Grant 62277046; in part by the Double First-Class Project of China University of Mining and Technology for Independent Innovation and Social Service under Grant 2022ZZCX06; and in part by the Six Talent Peaks Project in Jiangsu Province under Grant 2015-DZXX-010 and Grant 2018-XYDXX-044. 
{
    \small
    \bibliographystyle{ieeenat_fullname}
    \bibliography{main}
}
\clearpage
\setcounter{page}{1}
\setcounter{section}{0}
\setcounter{table}{0}
\setcounter{figure}{0}
\setcounter{equation}{0}

\maketitlesupplementary

\section{Oriented box$\to$2D Gaussian}
Different scaling factors $m$ will transform an oriented box into different scaling distributions, as shown in Fig~\ref{distribution_box_lambda}. 
For example, if approximately 99.7\% of the samples of a Gaussian distribution lie within a box, then six standard deviations correspond to the widths or heights according to the 68-95-99.7 rule, making $m=6$.
\distributionbox
\section{Reverse Process}
\textbf{\textit{Time 1$<$t$<$T.}} Inspired by~\cite{diffusiondet}, DDIM sampling strategy~\cite{ddim} is adopted, which allows for much faster sampling. When $1<t<T$, the reverse process is defined as:
\begin{equation}
  p_\theta(\bm{z}_{t-1}|\bm{z}_t):=\mathcal{N}(\bm{z}_{t-1};\boldsymbol{\mu}_{\theta}(\bm{z}_t,t),\boldsymbol{\Sigma}_{\theta}(\bm{z}_t,t))
\end{equation}
where
\begin{align}
  \boldsymbol{\mu}_\theta(\bm{z}_t,t)&=\sqrt{\bar{\alpha}_{t-1}}\left ( \frac{\bm{z}_t-\sqrt{1-\bar{\alpha}_t} \epsilon_\theta(\bm{z}_t,t)}{\sqrt{\bar{\alpha}_{t}}} \right)\nonumber\\
  &+\sqrt{1-\bar{\alpha}_{t-1}-\sigma^2_t}\epsilon_\theta(\bm{z}_t,t)\label{ddim_mean}\\
  \boldsymbol{\Sigma}_\theta(\bm{z}_t,t)&=\sigma ^2_t\mathbf{I}
\end{align} 
Formally, let $\mathbf{R}$ be the rotation transformation where $\mathbf{R}\in SO(2)$, we have:

\noindent
\textbf{\textit{Proposition 2.}} \textit{During $1<t<T$ in the reverse process $p_\theta$, the function $p_\theta(\bm{z}_{t-1}|\bm{z}_t)$ is rotation equivariant, i.e., $p_\theta(\bm{z}_{t-1}|\bm{z}_t)=p_\theta(\mathbf{R}(\bm{z}_{t-1})|\mathbf{R}(\bm{z}_t))$ under $SO(2)$, as long as $\mathbf{R}(\boldsymbol{\mu}_{\theta}(\bm{z}_t,t))=\boldsymbol{\mu}_{\theta}(\mathbf{R}(\bm{z}_t),t)$. Proof:}
\begin{align}
    &p_\theta(\mathbf{R}(\bm{z}_{t-1})|\mathbf{R}(\bm{z}_t))\nonumber\\
    &= \frac{1}{2\pi\sqrt{\left | \boldsymbol{\Sigma}_\theta(\mathbf{R}(\bm{z}_t),t)\right|}} \mathrm{exp} \{-\frac{1}{2}(\mathbf{R}(\bm{z}_{t-1})-\boldsymbol{\mu}_\theta(\mathbf{R}(\bm{z}_t),t)) ^\top\nonumber\\  
    &\quad\cdot\boldsymbol{\Sigma}_\theta(\mathbf{R}(\bm{z}_t),t)^{-1}\cdot(\mathbf{R}(\bm{z}_{t-1})-\boldsymbol{\mu}_\theta(\mathbf{R}(\bm{z}_t),t))\}\nonumber\\
    &= (2\pi\sqrt{\left |\sigma ^2_t\mathbf{I} \right|})^{-1} \mathrm{exp} \{-\frac{1}{2}(\mathbf{R}(\bm{z}_{t-1})-\mathbf{R}(\boldsymbol{\mu}_\theta(\bm{z}_t,t))) ^\top\nonumber\\  
    &\quad\cdot (1/\sigma^2_t)\mathbf{I} \cdot(\mathbf{R}(\bm{z}_{t-1})-\mathbf{R}(\boldsymbol{\mu}_\theta(\bm{z}_t,t)))\}\nonumber\\
    &= (2\pi\sqrt{\left |\sigma ^2_t\mathbf{I} \right|})^{-1} \mathrm{exp} \{-\frac{1}{2}(\mathbf{R}(\bm{z}_{t-1}-\boldsymbol{\mu}_\theta(\bm{z}_t,t))) ^\top\nonumber\\  
    &\quad\cdot (1/\sigma^2_t)\mathbf{I} \cdot\mathbf{R}(\bm{z}_{t-1}-\boldsymbol{\mu}_\theta(\bm{z}_t,t))\}\nonumber\\
    &= (2\pi\sqrt{\left |\sigma ^2_t\mathbf{I} \right|})^{-1} \mathrm{exp} \{-\frac{1}{2}(\bm{z}_{t-1}-\boldsymbol{\mu}_\theta(\bm{z}_t,t))^\top\mathbf{R}^\top\nonumber\\  
    &\quad\cdot (1/\sigma^2_t)\mathbf{I} \cdot\mathbf{R}(\bm{z}_{t-1}-\boldsymbol{\mu}_\theta(\bm{z}_t,t))\}\nonumber\\
    &= (2\pi\sqrt{\left |\sigma ^2_t\mathbf{I} \right|})^{-1} \mathrm{exp} \{-\frac{1}{2}(\bm{z}_{t-1}-\boldsymbol{\mu}_\theta(\bm{z}_t,t))^\top\nonumber\\  
    &\quad\cdot (1/\sigma^2_t)\mathbf{I} \cdot(\bm{z}_{t-1}-\boldsymbol{\mu}_\theta(\bm{z}_t,t))\}\nonumber\\
    &=p_\theta(\bm{z}_{t-1}|\bm{z}_t)
\end{align}

\textbf{\textit{Time t=1.}} The $p(\bm{z}_0)$ can be derived from the previous state. The $p(\bm{z}_0)$ is calculated as:
\begin{equation}
  \small{p_\theta(\bm{z}_0)=\hspace{-0.5mm}\int\hspace{-0.5mm} p_\theta(\bm{z}_{0:T})\mathrm{d} \bm{z}_{1:T}=\hspace{-0.5mm}\int \hspace{-0.5mm} p(\bm{z}_T)\prod_{t=1}^{T}p_\theta(\bm{z}_{t-1}|\bm{z}_t)\mathrm{d} \bm{z}_{1:T}}
\end{equation}
Formally, let $\mathbf{R}$ be the rotation transformation where $\mathbf{R}\in SO(2)$, we have:

\noindent
\textbf{\textit{Proposition 3.}} \textit{At time $t=1$ in the reverse process $p_\theta$, given the rotation invariant $p(\bm{z}_T)$, i.e., $p(\bm{z}_T)=p(\mathbf{R}(\bm{z}_T))$, and the rotation equivariant $p_\theta(\bm{z}_{t-1}|\bm{z}_t)$, i.e., $p_\theta(\bm{z}_{t-1}|\bm{z}_t)=p_\theta(\mathbf{R}(\bm{z}_{t-1})|\mathbf{R}(\bm{z}_t))$, then the $p_\theta(\bm{z}_0)$ is also rotation invariant, i.e., $p_\theta(\bm{z}_0)=p_\theta(\mathbf{R}(\bm{z}_0))$ under $SO(2)$.} Because:
\begin{align}
    p_\theta(\mathbf{R}(\bm{z}_0))&=\hspace{-0.5mm}\int\hspace{-0.5mm} p(\mathbf{R}(\bm{z}_T)) {\textstyle \prod_{t=1}^{T}} p_\theta(\mathbf{R}(\bm{z}_{t-1})|\mathbf{R}(\bm{z}_t))\mathrm{d} \bm{z}_{1:T}\nonumber\\
    &=\hspace{-0.5mm}\int \hspace{-0.5mm} p(\bm{z}_T){\textstyle \prod_{t=1}^{T}} p_\theta(\bm{z}_{t-1}|\bm{z}_t)\mathrm{d} \bm{z}_{1:T}\nonumber\\
    &=p_\theta(\bm{z}_0)
\end{align}

\section{Training loss}
In oriented object detection, there are two subtasks, classification for categories and regression for positions of objects. The losses $\mathcal{L}$ consist of the Focal loss $\mathcal{L}_{cls}$, $\ell_1$ loss $\mathcal{L}_{L1}$ and rotate IoU loss $\mathcal{L}_{riou}$:
\begin{equation}
  \mathcal{L}=\lambda_{cls}\mathcal{L}_{cls}+\lambda_{L1}\mathcal{L}_{L1}+\lambda_{riou}\mathcal{L}_{riou},
\end{equation}
where $\mathcal{L}_{cls}$, $\mathcal{L}_{L1}$, and $\mathcal{L}_{riou}$ are coefficient of corresponding losses. We adopt $\mathcal{L}_{cls}=2.0$, $\mathcal{L}_{L1}=2.0$, and $\mathcal{L}_{riou}=5.0$.
\section{Training configuration}
The training configuration is in Tab.~\ref{expsetting}.
\TableExpSetting
\section{Dataset}
\textbf{DOTA-v1.0} has 15 common categories: plane (PL), baseball diamond (BD), bridge (BR), ground track field (GTF), small vehicle (SV), large vehicle (LV), ship (SH), tennis court (TC), basketball court (BC), storage tank (ST), soccer-ball field (SBF), roundabout (RA), harbor (HA), swimming pool (SP), and helicopter (HC).

\textbf{DIOR-R} has 20 common categories: airplane (APL), airport (APO), baseball field (BF), basketball court (BC), bridge (BR), chimney (CH), expressway service area (ESA), expressway toll station (ETS), dam (DAM), golf field (GF), ground track field (GTF), harbor (HA), overpass (OP), ship (SH), stadium (STA), storage tank (STO), tennis court (TC), train station (TS), vehicle (VE), and windmill (WM).
\section{Nomenclature}
To facilitate clarity, we present a summary of symbols along with their corresponding descriptions as utilized in this study, encapsulated in Tab.~\ref{notations}.
\notations
\section{Main results}
\textbf{Results on DIOR-R.} The detailed results of every category on the DIOR-R are reported in Tab~\ref{tab:diorrappendix}.
\DIORRResultsappendix


\end{document}